\documentclass[10pt,journal,compsoc]{IEEEtran}
\pdfoutput=1
\usepackage{amsmath,amsfonts}
\usepackage{algorithmic}
\usepackage{algorithm}
\usepackage{array}
\usepackage[caption=false,font=normalsize,labelfont=sf,textfont=sf]{subfig}
\usepackage{textcomp}
\usepackage{stfloats}
\usepackage{url}
\usepackage{verbatim}
\usepackage{graphicx}

\usepackage{multirow}
\usepackage{booktabs}
\usepackage{chngcntr}
\usepackage{colortbl}
\usepackage{ulem} 
\usepackage{hyperref}
\usepackage{threeparttable} 
\usepackage{caption}       
\definecolor{mycolor}{RGB}{176,36,24}
\usepackage{xparse}
\usepackage{etoolbox}
\usepackage{xcolor}
\usepackage{ragged2e}  
\usepackage{titlesec}

\usepackage{svg} 
\newcolumntype{P}[1]{>{\centering\arraybackslash}p{#1}}
\usepackage{bm}
\usepackage{tabularx}
\usepackage[T1]{fontenc}
\usepackage{soul, color, xcolor}

\ExplSyntaxOn
\clist_new:N \l_my_color_bibkeys_clist
\clist_set:Nn \l_my_color_bibkeys_clist 
{jurcak200710}

\NewDocumentCommand{\instringTF}{m m m}
{
  \clist_if_in:NnTF \l_my_color_bibkeys_clist {#1}
    {#2} 
    {#3} 
}
\ExplSyntaxOff

\let\originalbibitem\bibitem

\renewcommand{\bibitem}[1]{%
    \instringTF{#1}
      {\color{red}\originalbibitem{#1}}
      {\color{black}\originalbibitem{#1}}
}

\ifCLASSOPTIONcompsoc
  \usepackage[nocompress]{cite}
\else
  \usepackage{cite}
\fi
\ifCLASSINFOpdf
\else
\fi
\begin{document}
%

\title{MPFNet: A Multi-Prior Fusion Network with a Progressive Training Strategy for Micro-Expression Recognition}
%
%
%
%

\author{
Chuang Ma, 
Shaokai Zhao, 
Dongdong Zhou, 
Yu Pei, 
Zhiguo Luo, 
Liang Xie,
Ye Yan\textsuperscript{\#},
Erwei Yin\textsuperscript{\#}
\IEEEcompsocitemizethanks{
\IEEEcompsocthanksitem E. Yin (yinerwei1985@gmail.com) and Y. Yan (yy\_taiic@163.com) are the corresponding authors. 
\IEEEcompsocthanksitem C. Ma, S. Zhao, Y. Pei, Z. Luo, L. Xie, Y. Yan, and E. Yin are with the Defense Innovation Institute, Academy of Military Sciences (AMS) and Intelligent Game and Decision Laboratory, Beijing, China.
\IEEEcompsocthanksitem Dongdong Zhou is with the School of Computer Science and Technology, Dalian University of Technology, Dalian, China.
}
}

\markboth{}%
{}

\IEEEtitleabstractindextext{%
\begin{abstract}
\justifying
Micro-expression recognition (MER), a critical subfield of affective computing, presents greater challenges than macro-expression recognition due to its brief duration and low intensity. While incorporating prior knowledge has been shown to enhance MER performance, existing methods predominantly rely on simplistic, singular sources of prior knowledge, failing to fully exploit multi-source information. This paper introduces the Multi-Prior Fusion Network (MPFNet), leveraging a progressive training strategy to optimize MER tasks. We propose two complementary encoders: the Generic Feature Encoder (GFE) and the Advanced Feature Encoder (AFE), both based on Inflated 3D ConvNets (I3D) with Coordinate Attention (CA) mechanisms, to improve the model’s ability to capture spatiotemporal and channel-specific features. Inspired by developmental psychology, we present two variants of MPFNet—MPFNet-P and MPFNet-C—corresponding to two fundamental modes of infant cognitive development: parallel and hierarchical processing. These variants enable the evaluation of different strategies for integrating prior knowledge. Extensive experiments demonstrate that MPFNet significantly improves MER accuracy while maintaining balanced performance across categories, achieving accuracies of 0.811, 0.924, and 0.857 on the SMIC, CASME II, and SAMM datasets, respectively. To the best of our knowledge, our approach achieves state-of-the-art performance on the SMIC and SAMM datasets.
\end{abstract}

\begin{IEEEkeywords}
micro-expression, prior learning, progressive training, meta-learning, attention mechanisms.
\end{IEEEkeywords}}

\maketitle

\IEEEdisplaynontitleabstractindextext

%
\IEEEpeerreviewmaketitle

\section{Introduction}
\IEEEPARstart{F}{acial} expressions play an essential role in conveying human emotions and reflecting psychological states during interpersonal interactions. Micro-expressions (MEs) are involuntary facial expressions that often occur when individuals attempt to suppress or conceal their true emotions, making them crucial for uncovering genuine emotional states \cite{xie2022overview}. Micro-expression recognition (MER) has numerous potential applications, including mental health monitoring, human-computer interaction, and security enforcement. By detecting subtle emotional cues, MER offers valuable insights that can inform decision-making, which has led to increasing interest in the field in recent years\cite{wu2010micro}.

\begin{figure}[t]
    \centering
    \includegraphics[width=1\linewidth]{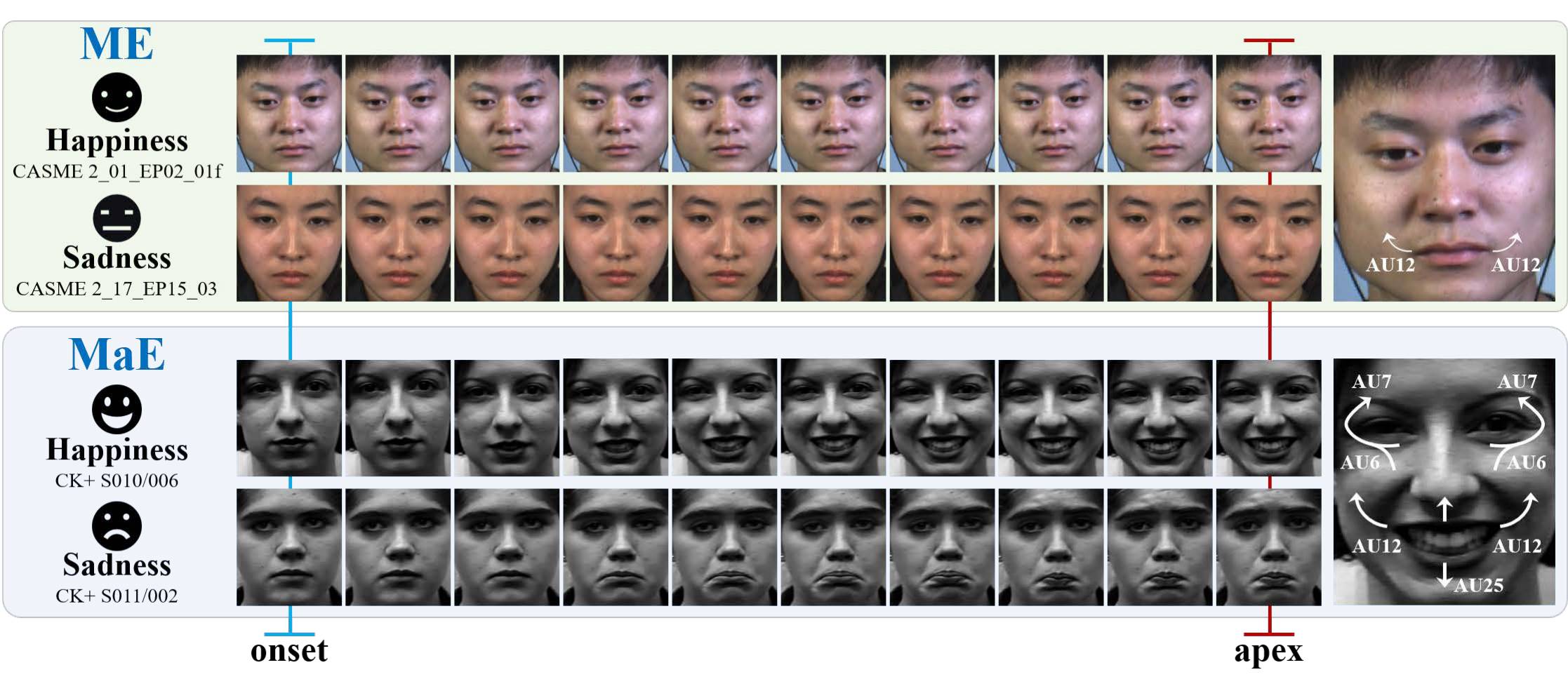}
    \caption{Comparison of MEs and MaEs. The top and bottom rows show examples of happiness and sadness from the CASME II and CK+ dataset, respectively. White arrows indicate the muscle movement direction of the activated facial action units.}
    \label{figs:ME_MaE}
\end{figure}

Compared to the more apparent macro-expressions (MaEs) observed in daily life, MEs, although also grounded in Ekman's basic emotion model \cite{tracy2011four} (e.g., happiness, anger, sadness, surprise, fear, and disgust), exhibit notable differences. First, in terms of appearance, MEs involve subtle and rapid muscle movements localized to specific facial regions, typically lasting only between 1/25 and 1/3 of a second \cite{zhao2023facial}. These characteristics make the detection and recognition of MEs considerably more challenging than MaEs \cite{li2022deep} (see Fig. \ref{figs:ME_MaE}). Second, in facial expression coding, although MEs share similarities with MaEs, their transient and subtle nature necessitates specialized expertise and detailed manual annotation during analysis. Psychologists often use the Facial Action Coding System (FACS) \cite{ekman1978facial} to analyze MEs, but this process is both time-consuming and labor-intensive. Finally, in terms of feature representation learning, while many models successful in MaE recognition—such as Convolutional Neural Networks (CNN) \cite{zhao2021two,li2019micro,wang2024htnet}, Recurrent Neural Networks (RNN) \cite{xia2019spatiotemporal,zhang2025towards}, and Transformers \cite{zhang2022short,li2024micro}—are beginning to be applied to the ME domain, the fleeting, localized, and subtle nature of MEs presents additional challenges for these models. Addressing these challenges requires particular focus on three key areas: optimizing the model’s learning process, tackling issues of data sparsity and imbalance, and efficiently extracting fine-grained local features.

Regarding model optimization, the complexity and subtlety of MEs make efficient recognition challenging when relying solely on data-driven feature learning. The integration of prior knowledge has been identified as an effective strategy to optimize the learning process and improve model performance. For instance, research has demonstrated that leveraging facial micro-movement patterns \cite{allaert2019micro}, the relationships between different Action Units (AUs) \cite{lei2021micro}, and transferring knowledge from MaEs \cite{sun2020dynamic} can enhance the model's ability to identify and learn ME features. To address the issues of sample sparsity and imbalance, researchers have employed various data augmentation strategies \cite{li2020local}, such as rotating, scaling, flipping, or generating new samples, to expand the dataset and alleviate sample imbalance. Transfer learning techniques have also been applied, transferring knowledge from larger datasets related to similar tasks, which improves model adaptability in the target domain \cite{tang2024facial, gan2024transfer}. Furthermore, meta-learning approaches provide innovative solutions for overcoming data limitations \cite{wang2025cross, gong2023meta}. In terms of feature extraction, CNNs have been widely used in MER tasks \cite{zhou2023coutfitgan, zhou2024learning}. However, due to the subtle muscle movements involved in MEs, CNNs still face limitations in capturing such fine-grained details. To address this issue, recent studies have introduced various attention mechanisms \cite{guo2022attention,cai2024mfdan,wu2024micro}, which dynamically adjust weights and integrate global information to improve model adaptability and accuracy.

Despite these advancements, MER still faces numerous challenges. For example, the prior knowledge used in model optimization is often overly simplistic or incomplete, preventing the model from fully realizing its potential \cite{arpit2017closer,zhang2023facial}. Data augmentation methods, such as sample synthesis, may introduce misleading artifacts into the model \cite{uchinoura2024improved}, negatively impacting its generalization ability. Furthermore, due to significant domain differences between ME and MaE, the effectiveness of transfer learning may be limited \cite{weiss2017comparing, zong2018domain}. In feature extraction, although attention mechanisms have been applied, further improvements are needed to efficiently capture the subtle characteristics of MEs.

To address these challenges, we propose a Multi-Prior Fusion Network (MPFNet) based on a progressive training strategy, aiming to improve MER performance from three perspectives: model optimization, data processing, and feature extraction. Our approach is inspired by the multi-stage cognitive development process in human infants \cite{he2020multi}, where infants progressively deepen their understanding of the world through continuous interactions between domain-general learning mechanisms and evolving environmental experiences. In the early stages, infants learn basic object features by recognizing similarities and differences. As cognitive abilities mature, they can distinguish more complex features and perform more advanced classification tasks. This simple-to-complex cognitive progression provides inspiration for our model design. Specifically, we first develop a triplet network designed to minimize the feature distance within the same category while maximizing the distance between features from different categories in the embedding space during training. This contrastive learning strategy generates an encoder capable of effectively extracting general ME features. Subsequently, we utilize a self-constructed motion-enhanced, sample-balanced ME dataset to train an advanced feature encoder, enabling the capture of more complex ME features.

To optimize the synergy between these two encoders, we design two architectures: MPFNet-P with parallel feature encoders and MPFNet-C with cascaded feature encoders. These architectures implement distinct mechanisms for integrating prior knowledge, corresponding to two fundamental modes of infant cognitive development: parallel processing and hierarchical processing. This design is grounded in established theories from developmental psychology. Specifically, Lewkowicz \textit{et al.} \cite{lewkowicz2009emergence} argued that infant cognition does not follow a linear processing pattern but instead involves simultaneous engagement with multiple dimensions of information without strict hierarchical prioritization. Correspondingly, Cohen \textit{et al.} \cite{cohen2002constructivist} proposed a hierarchical cognitive development theory, suggesting that infants' learning systems exhibit a structured hierarchy, where the ability to process complex information is progressively built upon lower-level processing capabilities. Building on these theoretical foundations, the parallel architecture of MPFNet-P simulates the infant cognitive mode of synchronous multisensory integration, whereas the cascaded architecture of MPFNet-C reflects the hierarchical information processing mode, where higher-level features are progressively constructed upon lower-level representations. Both the MPFNet-P and MPFNet-C architectures employ the Inflated 3D Convolutional Networks (I3D) \cite{carreira2017quo} as their encoder backbone, augmented with Coordinate Attention (CA) blocks \cite{hou2021coordinate}. This integration, termed the CA-I3D model, enhances the network's ability to extract meaningful spatiotemporal features by performing 3D convolutions across consecutive frames.

Furthermore, we introduce meta-learning to simulate the rapid adaptability observed in infants as they learn different tasks. Meta-learning trains the model across multiple tasks, enabling it to efficiently extract prior knowledge for adapting to new ones. Finally, to evaluate MPFNet's performance, we conduct experiments on several publicly ME datasets. The results show that both MPFNet-P and MPFNet-C outperform the baseline model in MER tasks, with MPFNet-C achieving particularly strong results. This demonstrates that our progressive training strategy effectively integrates multi-level prior knowledge, enhancing the model's ability to classify MEs.

To sum up, the main contributions of this research are:

        \begin{enumerate}
        \item We propose a novel MPFNet to extract both generic and advanced features of MEs, leveraging complementary prior knowledge to enhance MER. MPFNet comprises two variants: MPFNet-P and MPFNet-C, which explore prior fusion strategies from feature diversity and hierarchy perspectives.
        \item To address the challenges of limited samples and class imbalance in ME datasets, we propose a data augmentation method using dynamic motion magnification. To capture critical spatiotemporal information of MEs, we utilize the CA-I3D model as the backbone for feature encoder, integrating an attention mechanism into the I3D framework to effectively model channel relationships and long-term dependencies.
        \item Extensive experiments and visual analyses demonstrate that our approach overcomes the limitations of traditional single-prior knowledge. By integrating multiple complementary priors, we significantly enhance model performance, improving overall classification accuracy while ensuring balanced results across categories, thus achieving competitive performance.
	    \end{enumerate}

The remainder of this paper is structured as follows. Section \ref{related_work} reviews related work, followed by the details of MPFNet in Section \ref{method}. Section \ref{experiments} introduces experimental data, evaluation metrics and implementation details. Section \ref{Results and analysis} presents the experimental results, ablation study and visualization analysis. Section \ref{conclusion} concludes the paper. Finally, Section \ref{Ethical} discusses the ethical issues related to this research.

\section{Related work}
\label{related_work}

This study focuses on three key areas of MER: model learning, data augmentation, and feature extraction. In this section, we first provide a brief review of related studies in these areas. Building on this foundation, we then analyze the unique characteristics and innovations of our work, highlighting its advantages over existing approaches.

\subsection{Prior learning strategies in ME analysis models} 

Studies have shown that incorporating prior knowledge into deep learning models can effectively guide them to focus on critical features of MEs, achieving superior performance in tasks such as ME recognition, spotting, and generation. For MER, Sun \textit{et al.} \cite{sun2020dynamic} distilled and transferred knowledge from facial action units (AUs), using features from the teacher network as prior knowledge to guide the student part to effectively learn from the target ME dataset. Additionally, Wei \textit{et al.} \cite{wei2023prior} proposed a decomposition and reconstruction graph representation learning model, integrating prior knowledge of the relationship between different AUs, improving the model's interpretability and feature learning capabilities. For ME spotting, Yin \textit{et al.} \cite{yin2023aware} encoded prior knowledge about motion patterns of MEs into the network, improving spatial feature embedding and alleviating over-fitting. Besides, Wang \textit{et al.} \cite{wang2021mesnet} also tried to solve ME spotting through convolutional neural networks and constrained the network's complexity by introducing additional distribution prior knowledge. This approach helps alleviate the overfitting problem in ME detection. In the field of ME generation, several studies have demonstrated that utilizing prior knowledge of MEs can significantly improve the quality of reconstructed videos. For instance, Zhang \textit{et al.} \cite{zhang2023facial} proposed a facial prior-guided ME generation framework, which utilized a facial prior module to guide the motion representation and generation of ME, significantly improving the performance of the ME generation model.
Although prior learning strategies have been applied in ME analysis, the prior knowledge utilized in existing studies remains relatively simple and lacks the integration of multi-source information. In this study, we utilize the capabilities to learn both generic and advanced features as complementary prior knowledge to assist in MER. To the best of our knowledge, no existing studies in the literature have adopted the concept of multi-prior fusion for MER.

\begin{figure*}[t]
    \centering
    \includegraphics[width=0.9\linewidth]{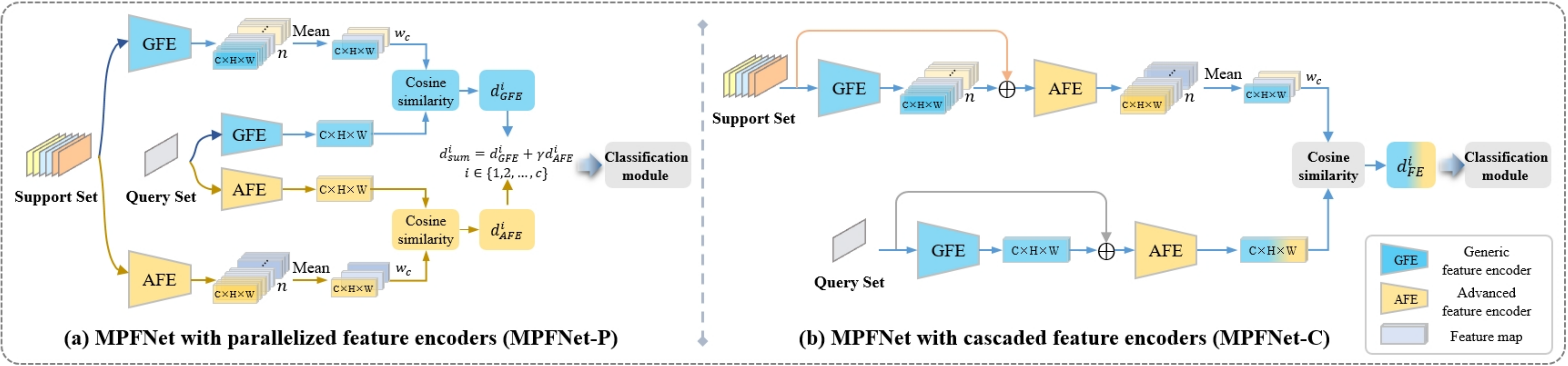}
    \caption{The overall architecture of the proposed MPFNet, which comprises two distinct variants: MPFNet-P and MPFNet-C. Both variants incorporate two feature encoders—the Generic Feature Encoder (GFE) and the Advanced Feature Encoder (AFE). MPFNet-P employs a parallel encoder architecture, whereas MPFNet-C utilizes a cascaded encoder structure, each optimized for feature extraction through their respective configurations.}
    \label{figs:MPFNet}
\end{figure*}

\subsection{Data enhancement methods to tackle few-shot and imbalanced data} 

Addressing the challenges of few-shot learning and class imbalance in MER has been a significant focus of recent research. Various approaches have been proposed to tackle these issues effectively, including data augmentation, transfer learning, and meta-learning. For example, Xia \textit{et al.} \cite{xia2019spatiotemporal} employed temporal data augmentation strategies to enhance the limited training samples and utilized a balanced loss function to address the issue of imbalanced training. Subsequently, Xie \textit{et al.} \cite{xie2020assisted} proposed a data augmentation method to generate ME images using the action units intensity extracted from MEs as training conditions to alleviate the limited and unbalanced problem of existing MER datasets. Additionally, transfer learning is another widely used technique, leveraging knowledge from related tasks or larger datasets to improve performance of MER. Xia \textit{et al.} \cite{xia2020learning} proposed a MER framework that leverages MaE samples for guidance and employs an adversarial learning strategy and triplet loss to capture the shared features of ME and MaE samples. More recently, to address the issue of insufficient data for MER, Tang \textit{et al.} \cite{tang2024facial} proposed a dual graph convolutional network architecture with transfer learning. However, augmented data poses a risk of overfitting due to its similarity to the original data. Synthetic samples can introduce artificial artifacts that absent in the natural world, potentially misleading the model. Moreover, substantial domain discrepancy between MEs and MaEs may hinder the effectiveness of transfer learning. To address these challenges, this study proposes a data augmentation method based on dynamic motion magnification. This approach not only enhances the intensity of subtle movements in ME videos but also dynamically adjusts the augmentation effect based on the distribution of the original samples. As a result, it effectively mitigates issues of limited sample size and imbalanced class distribution in ME datasets.

\subsection{Feature extraction methods for MER}

Due to the subtle and difficult-to-discern movements of facial muscles in MEs, the effectiveness of MER largely depends on the discriminative features. Recent studies have primarily focused on leveraging high-level features derived from deep learning models. For instance, Zhao \textit{et al.} \cite{zhao2021two} employed 3D convolutional neural networks (3D-CNNs) to encode both spatial and temporal information, aiming to capture comprehensive representations, including motion cues and long-sequence dependencies. Similarly, Thuseethan \textit{et al.} \cite{thuseethan2023deep3dcann} used 3D-CNNs to learn useful spatiotemporal features from facial images and then combined the learned features and the semantic relationships between the regions to predict the MEs. However, traditional 3D-CNNs demand substantial parameters and computational resources, which constrains their ability to capture the subtle variations in MEs. Recently, the attention mechanism enables models to concentrate on the most pertinent aspects of input data, leading to its extensive adoption in the field of MEs research. For example, Zhou \textit{et al.} \cite{zhou2023inceptr} proposed a dual-branch attention network for MER, which adopts the convolutional block attention module (CBAM) to enable the model to capture the most discriminative multi-scale local and global features. Shu \textit{et al.} \cite{shu2023res} incorporated a squeeze-excitation (SE) block into the network. This SE block highlights valuable ME features while suppressing irrelevant ones. More recently, Liong \textit{et al.} \cite{liong2024sfamnet} proposed a multi-stream MER network based on attention mechanism, which can predict recognition confidence scores and emotion labels. The above studies demonstrate the effectiveness of the attention mechanism for MER. To capture essential temporal, spatial, and channel-specific features for MER, we integrate the CA block with the I3D model. The CA block captures cross-channel, direction-aware, and position-sensitive information, while the I3D model excels at extracting robust spatiotemporal features, facilitating comprehensive modeling of temporal dynamics in ME videos. This integration enhances the model's ability to detect key features and significantly improves its sensitivity to subtle variations in MEs, resulting in a more accurate and robust feature representation for MER.

\vspace{-0.1cm}
\section{Methodology}
\label{method}

Human beings can acquire new skills with just a few examples and learn even faster when faced with novel, related tasks. This ability is attributed to the human capacity to learn and utilize various forms of prior knowledge \cite{maguire1999functional}. Related studies have also leveraged prior knowledge from other domains through pre-trained deep neural networks \cite{wei2023prior, yin2023aware}. Inspired by this, we propose a novel multi-prior fusion network (MPFNet) for MER based on a progressive training strategy. This Section begins with an overview of the proposed MPFNet architecture in Section \ref{sec: overview}. Section \ref{sec: Data preprocessing} describes the preprocessing steps for the input data. In Section \ref{sec: Pre-training}, we provide a detailed explanation of the pretraining process of the feature encoder. Each stage is specifically designed to correspond to the learning process of different types of prior knowledge. To investigate effective strategies for integrating multiple types of prior knowledge, we introduce two model variants, MPFNet-P and MPFNet-C, with their operational mechanisms analyzed in Section \ref{sec: variants}. Finally, Section \ref{sec: classification} discusses the design and implementation of the classification module, focusing on the classifier architecture and the selection of the loss function.

\subsection{Overview of the MPFNet architecture}
\label{sec: overview}

The overall architecture of MPFNet is illustrated in Fig. \ref{figs:MPFNet}. Operating within a metric-based meta-learning framework, MPFNet processes both the support set and the query set as inputs and outputs classification results for the query set. MPFNet offers two model variants: MPFNet-P and MPFNet-C. Both variants consist of two feature encoders: the Generic Feature Encoder (GFE) and the Advanced Feature Encoder (AFE). These encoders are designed to capture prior knowledge at distinct levels of abstraction, thereby enhancing feature representation and classification performance. MPFNet-P employs a parallel encoder architecture, facilitating the model to extract features from multiple perspectives and capture more diverse information. In contrast, MPFNet-C utilizes a cascaded encoder architecture, which progressively refines feature representations to capture subtle and discriminative characteristics. This stepwise refinement allows the model to gradually focus on the subtle yet critical features embedded in MEs, further enhancing recognition performance. By comparing these two variants, we aim to evaluate the impact and effectiveness of different prior knowledge integration strategies on MER.


\subsection{Data preprocessing}
\label{sec: Data preprocessing}

\begin{figure}[t]
        \centering
        \includegraphics[width=0.93\linewidth]{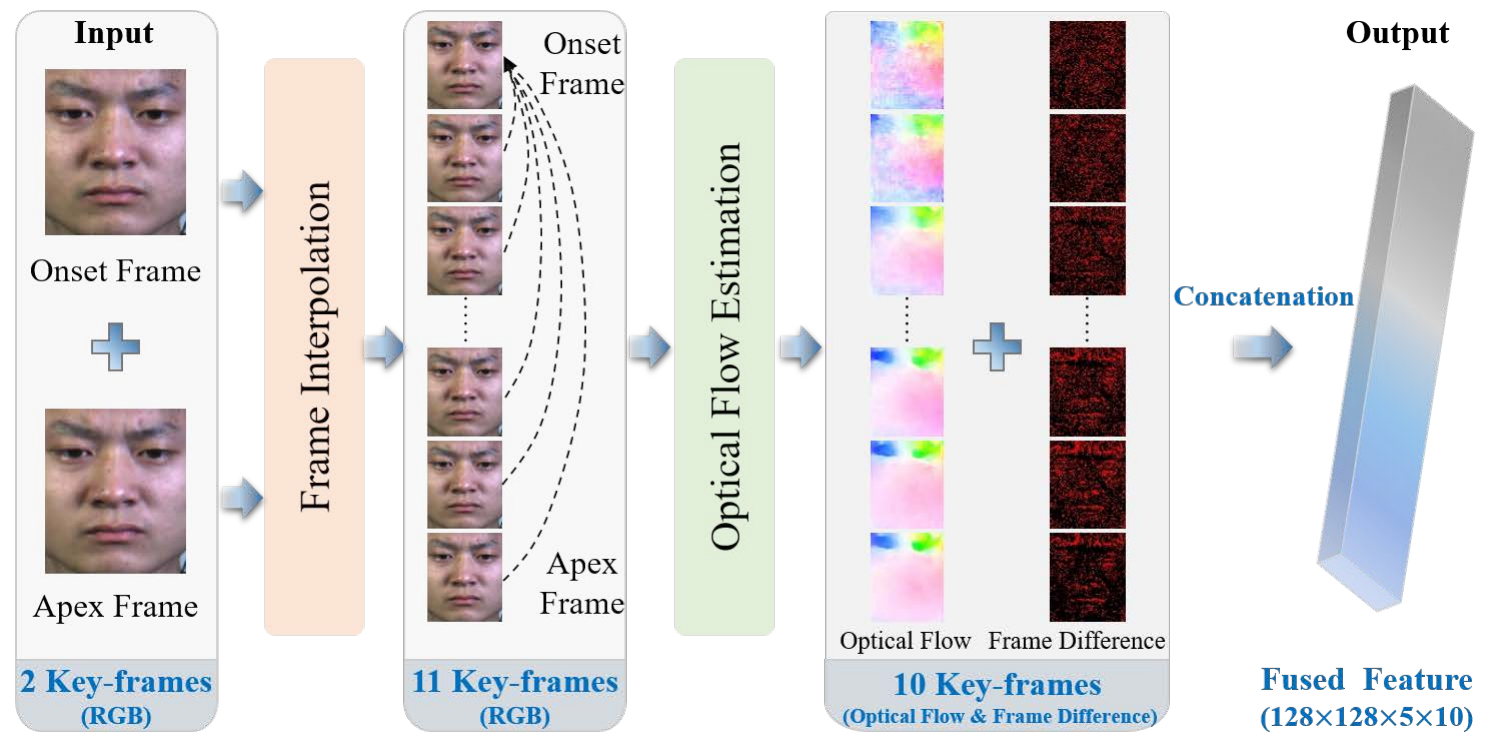}
        \caption{Data preprocessing steps, including the generation of ME frames with normalized length using VFI, followed by the computation and integration of optical flow and frame difference features.}
        \label{figs:Feature_fuse}
    \end{figure}

In this section, we implement a comprehensive preprocessing pipeline to extract robust facial features. First, we employ the face recognition algorithm provided by Alibaba Cloud\footnote{url{https://vision.aliyun.com/facebody}} to perform precise face detection, alignment, and cropping on ME images, minimizing the interference from non-facial areas and head pose variations. The cropped facial regions are resized to 128$\times$128 pixels. Subsequently, to address the significant variability in ME sequence lengths and the inherent noise in high-speed camera recordings, we adopt a keyframe-based frame interpolation method for sequence normalization. Following this, we extract and integrate two complementary feature modalities—inter-frame optical flow and frame difference features—to comprehensively represent the spatiotemporal characteristics of MEs. The preprocessing workflow is illustrated in Fig. \ref{figs:Feature_fuse}.

\subsubsection{Frame interpolation method}

In ME analysis, the frame sequence from the onset frame to the apex frame effectively reflects the dynamic characteristics of facial muscle movements and their evolving trends. However, these sequences often exhibit inconsistent frame lengths, ranging from 9 to over 100 frames. For sequences with a higher frame count, direct downsampling may result in unsmooth motion trends and potential loss of critical motion information. Conversely, sequences with insufficient frames require effective upsampling strategies to reconstruct their motion information. To address these issues, we employs a keyframe-based video frame interpolation (VFI) algorithm, designed to achieve two primary objectives: (i) generating standardized-length frame sequences, and (ii) preserving the temporal dynamics of MEs, thereby producing smoother and more continuous motion patterns that are crucial for capturing subtle ME features. Specifically, we utilize the VFI model proposed by Zhang et al. \cite{zhang2023extracting}. This model innovatively restructures the information processing mechanism of inter-frame attention, enhancing appearance feature representation through attention maps and effectively capturing motion dynamics. In implementation, the onset frame $I_o$ and apex frame $I_a$ are used as inputs to the VFI model to produce fixed-length ME sequences of 11 frames. The interpolation process can be formulated as: 

    \begin{equation}
        I_t=VFI(I_o, I_a, t), \quad t \in \{1, 2, \ldots, 9\},
    \end{equation}
where $I_t$ represents the interpolated frame at time step $t$, and $VFI(\cdot)$ denotes the frame interpolation function.

The selection of 11 frames strikes a balance between capturing high-resolution motion features and maintaining computational efficiency. Furthermore, this length selection is informed by the successful practices of previous studies \cite{zhao2021two, guo2019extended, khor2018enriched, li2013spontaneous}. This approach not only preserves the spatiotemporal information of the original ME videos and eliminates redundant frames but also accentuates the motion dynamics of the peak frame, providing robust input data for subsequent feature extraction and classification. The impact of this hyperparameter on model performance is analyzed in Section \ref{Results and analysis}.

\subsubsection{Calculation and integration of optical flow and frame difference features}

    \begin{figure}[b]
        \centering
        \includegraphics[width=0.8\linewidth]{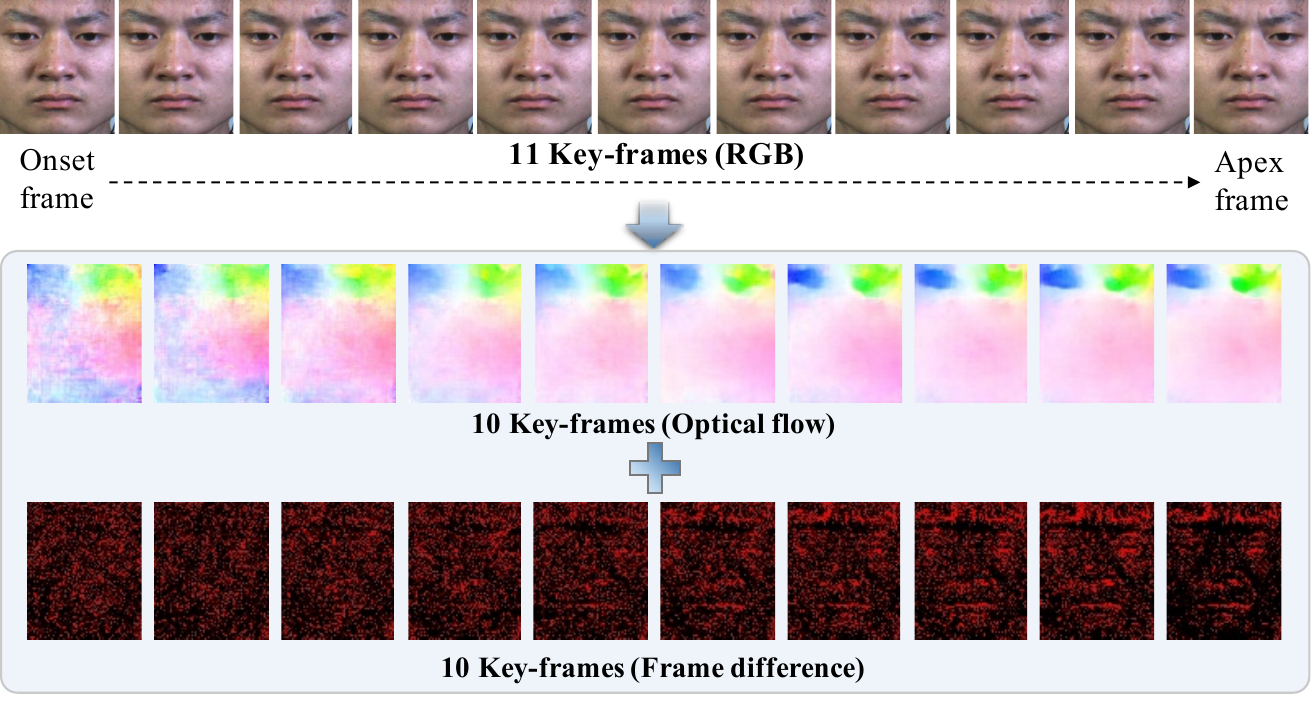}
        \caption{The process of obtaining optical flow and frame difference features. It can be seen that both features progressively become more prominent, effectively capturing the movement patterns of MEs.}
        \label{figs:OF_FD}
    \end{figure}

For the fixed-length ME frame sequences, we compute inter-frame optical flow and frame difference features as handcrafted inputs for the model. Optical flow features capture pixel-level motion between frames, while frame difference features represent the intensity variations of each pixel between consecutive frames. These features are crucial for capturing the subtle muscular movements essential for MER and are inherently complementary. In this study, we employ the deep learning-based optical flow estimation method, FlowNet 2.0\footnote{\url{https://github.com/NVIDIA/flownet2-pytorch}}, which has proven effective in detecting subtle motions in videos, to extract optical flow features. The optical flow between two consecutive frames $I_t$ and $I_{t+1}$ is computed as a displacement vector field $(u_t,v_t)$, where $u_t$ and $v_t$ represent the horizontal and vertical displacements, respectively. This can be expressed as:

    \begin{equation}
        \left(u_{t}, v_{t}\right)={FlowNet}\left(I_{t}, I_{t+1}\right), \quad t \in\{0,1, \ldots, 9\}.
    \end{equation}

The resulting optical flow features for the entire sequence are represented as a tensor $\mathcal{O} \in {R}^{128 \times 128 \times 2 \times 10}$, where 2 corresponds to the horizontal and vertical displacement components, and 10 denotes the number of inter-frame pairs.
The frame difference features are computed as the pixel-wise intensity difference between consecutive frames for each RGB channel. For a given channel $c \in \{R,G,B\}$, the frame difference $\Delta I_{t}^{c}$ at time $t$ is calculated as:

    \begin{equation}
        \Delta I_{t}^{c}=\left|I_{t+1}^{c}-I_{t}^{c}\right|, \quad t \in\{0,1, \ldots, 9\}.
    \end{equation}

The resulting frame difference features for the entire sequence are represented as a tensor $\mathcal{D} \in {R}^{128 \times 128 \times 3 \times 10}$, where 3 corresponds to the RGB channels.
To integrate the optical flow features $\mathcal{O}$ and frame difference features $\mathcal{D}$, we concatenate them along the channel dimension, resulting in a fused feature tensor $\mathcal{F} \in {R}^{128 \times 128 \times 5 \times 10}$:

    \begin{equation}
        \mathcal{F}={Concat}(\mathcal{O}, \mathcal{D}),
    \end{equation}
where $Concat(\cdot)$ denotes the concatenation operation along the channel dimension. This fused feature combines both motion dynamics (via optical flow) and intensity variations (via frame differences), providing a comprehensive representation of the ME sequence for subsequent analysis.
The extraction of optical flow and frame difference features is shown in Fig. \ref{figs:OF_FD}. Notably, as the ME progresses, both features become increasingly pronounced.

\subsection{Pre-training of feature encoders}
\label{sec: Pre-training}

    \begin{figure}[b]
        \centering
        \includegraphics[width=0.9\linewidth]{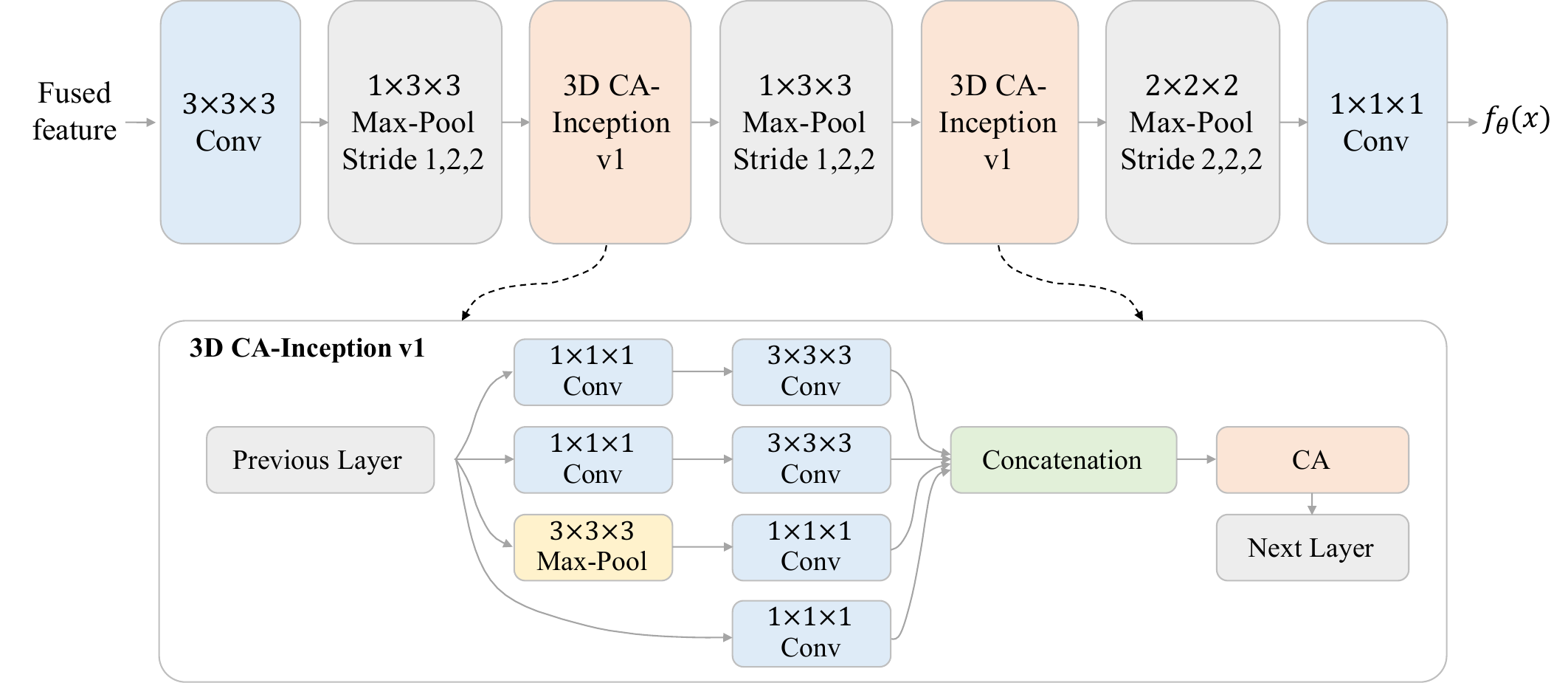}
        \caption{The structure of the proposed CA-I3D model. We optimize the original I3D network to make the model more suitable for MER task.}
        \label{figs:CA-I3D}
    \end{figure}

MPFNet comprises two feature encoders: the GFE and the AFE. The GFE is pretrained using a triplet network-based prior learning approach to extract general features for MER. Meanwhile, the AFE is pretrained on a larger, more balanced dataset derived from the original ME dataset using a motion amplification model, enabling it to capture advanced ME features. The prior knowledge acquired from both encoders is then utilized to initialize the model parameters as convolutional layer weights. Finally, the model is retrained on the original ME dataset, with encoder parameters fine-tuned to achieve accurate ME classification. Both the GFE and AFE utilize the CA-I3D architecture as their backbone, the details of which will be elaborated in the following section.

\subsubsection{CA-I3D}

In this study, we integrate the I3D architecture with CA Block to construct the backbone of our feature encoder, termed CA-I3D, for precise spatiotemporal feature extraction of MEs. The I3D model is an extension of 2D convolutional networks, which introduces a temporal dimension by extending traditional 2D convolutional and pooling kernels into 3D forms, thereby enabling the modeling of dynamic information within video sequences. Meanwhile, the CA Block enhances feature representation by effectively capturing channel relationships and long-range dependencies through precise positional information. The overall architecture of the proposed CA-I3D model is illustrated in Fig. \ref{figs:CA-I3D}. The CA-I3D model comprises multiple 3D convolutional layers, max-pooling layers, and 3D CA-Inception v1 modules \cite{szegedy2015going}. It takes the fused feature tensor $\mathcal{F}$ as input and produces the deep feature vector $f_{\theta}(x)$ for a given sample $x$, as defined below:

    \begin{equation}
        f_{\theta } (x)=\textit{CA-I3D}(\mathcal{F}).
    \end{equation}

To better align with the requirements of MER tasks, we optimized the original I3D network architecture. The model begins with a 3$\times$3$\times$3 convolutional layer for spatial feature extraction, followed by a 1$\times$3$\times$3 max-pooling layer with strides of (1, 2, 2), which performs pooling along the height and width dimensions while preserving the channel and temporal dimensions.  
The 3D CA-Inception v1 module employs multiple convolutional filters of varying sizes (e.g., 1$\times$1$\times$1 and 3$\times$3$\times$3) to capture diverse spatial patterns across different scales. It consists of multiple parallel convolutional branches to extract scale-specific features. The CA module is integrated after the concatenation layer of the 3D Inception v1 module within the I3D architecture. Another 1$\times$3$\times$3 max-pooling layer with strides of (1, 2, 2) is applied to downsample the feature map while preserving essential information. An additional 3D CA-Inception v1 module is then incorporated to further enhance the network’s ability to capture complex spatial patterns. The network concludes with a linear layer to facilitate nonlinear transformations.  
All convolutional layers use rectified linear unit (ReLU) activation, and network weights are initialized randomly following a standard normal distribution with a mean of 0 and a variance of 1. To prevent the loss of low-level image features typically associated with pooling operations, the first max-pooling layer was removed. Additionally, the final average pooling layer was eliminated, retaining only the convolutional layers. This modification not only reduces the number of parameters but also preserves global image information, enhancing the network's robustness. Furthermore, to mitigate the risk of overfitting, the number of Inception modules was reduced from nine to two.

\subsubsection{Pre-training of the GFE}

The pre-training of the GFE is achieved through prior learning based on a triplet network, enabling the GFE to acquire the capability of extracting general features for MER. This capability allows the model to effectively distinguish similarities and differences between samples of different categories. Based on the assumption that ME samples from the same category should form tight clusters in the embedding space, we constructed a triplet network to ensure that samples with the same label are closely positioned in the embedding space, while samples with different labels are positioned farther apart. The triplet network consists of three CA-I3D components that share the same architecture and parameters. The input of the network is a series of triple samples defined as $[x_a,x_p,x_n]$, which consists of an anchor sample $x_a$, a positive sample $x_p$ from the same category, and a negative sample $x_n$ from a different category. During the training process, for an input triplet sample, the model will output three feature vectors $f(x_a)$, $f(x_p)$ and $f(x_n)$ after the processing of CA-I3D modules. We employ a triplet loss function to learn discriminative feature embedding, such that the embedded distance of the positive pair (images of the same class) is closer than that of the negative pair (images of different classes) by a distance margin. The triplet loss is presented as follows:

	\begin{equation}
        \small
		\mathcal{}{L_{t}} = \sum_{x_{a}} \max \left (d\left(f\left(x_{a}\right), f\left(x_{p}\right)\right)^{2}-d\left(f\left(x_{a}\right), f\left(x_{n}\right)\right)^{2}+\alpha, 0 \right ),
	\end{equation}
where $\alpha$ is a hyperparameter that controls the margin between the distances of $x_p$ and $x_n$. We use the Euclidean distance between ME features as a metric, which is described as follows:

	\begin{equation}
		\mathcal{} d(f(x_{a}),f(x_{n}))=\left \| f(x_{a})-f(x_{n})\right \| _{2}.
	\end{equation}


After several epochs of training on the ME dataset, the network reached a fixed error over the triplet comparisons. Then we use the parameters of the convolutional layer of the trained CA-I3D model to initialize our GFE to extract the generic features of MEs.



\subsubsection{Pre-training of the AFE}

    \begin{table*}[t]
    \centering
    \large
        \caption{The process of sample equalization for the ME datasets used in this study}
            \resizebox{0.95\textwidth}{!}{
                \begin{tabular}{c  P{1.4cm} P{2.3cm} P{0.8cm} P{1.5cm} P{2.3cm} P{0.8cm} P{1.5cm} P{2.3cm} P{0.8cm} P{1.5cm} P{2.3cm}}
                    \hline
                    \toprule[0.5pt]
                    \multicolumn{1}{c}{\multirow{2}{*}{Category}} & \multicolumn{3}{c}{SMIC-HS \cite{pfister2011recognising}} & 
                    \multicolumn{3}{c}{CASME II \cite{yan2014casme}} & 
                    \multicolumn{3}{c}{SAMM \cite{davison2016samm}} & \multicolumn{2}{c}{MEGC2019-CD \cite{see2019megc}} 
                    \\ \cmidrule(r){2-4} \cmidrule(r){5-7} \cmidrule(r){8-10} \cmidrule(r){11-12}
                    \multicolumn{1}{c}{} & 
                    Raw-MEs & Amplified-MEs & $\varphi$ & 
                    Raw-MEs & Amplified-MEs & $\varphi$ & 
                    Raw-MEs & Amplified-MEs & $\varphi$ &
                    Raw-MEs & Amplified-MEs
                    \\ \hline
                    Negative & 70 & 630 & 1$\sim$8 & 88 & 440 & 1$\sim$4 & 92 & 368 & 1$\sim$3 & 250 & 1438 \\
                    Positive & 51 & 663 & 1$\sim$12 & 32 & 416 & 1$\sim$12 & 26 & 260 & 1$\sim$9 & 109 & 1339 \\
                    Surprise & 43 & 645 & 1$\sim$14 & 25 & 375 & 1$\sim$14 & 15 & 225 & 1$\sim$14 & 83 & 1245 \\
                    Total & 164 & 1938 & -- & 145 & 1231 & -- & 133 & 853 & -- & 442 & 4022 \\ \hline 
                    \toprule[0.5pt]
                \end{tabular}}
        \label{table:equalization}
    \end{table*}

The pretrained GFE is capable of extracting general features of MEs. However, relying solely on these general features is insufficient for accurately modeling the fine-grained information required for MER tasks. To address this limitation, we designed and pretrained an additional feature encoder, termed the AFE, to enhance the model’s ability to capture advanced features essential for MER. These advanced features include subtle facial movement patterns, complex texture characteristics, and fine-grained motion details. The pretraining of the AFE is based on a prior learning strategy that utilizes a balanced, motion-amplified ME dataset. Specifically, we employed the video magnification model proposed by Tae-Hyun \textit{et al.} \cite{oh2018learning}, using frame sequences from the onset to the apex frames of MEs as input. To address the imbalance in sample distribution across categories, we dynamically adjusted the magnification factors based on each category's sample size. Categories with smaller sample sizes were assigned a greater range of magnification factors, allowing for the generation of more synthetic samples and improving dataset balance. This approach not only enhances the representation of subtle motions in videos but also effectively mitigates sample imbalance within the dataset.
As defined by Wu et al. \cite{wu2012eulerian} in their work on motion magnification, a single frame within a continuous video can be described as:

	\begin{equation}
		\mathcal{}{I(x,t)} = f(x+\delta (x,t)),
	\end{equation}
where $\delta(x,t)$ is the motion field at position $x$ and time $t$.
By performing motion magnification on the original image, we can obtain image $I_{mag}$:

	\begin{equation}
		\mathcal{}{I_{mag}(x,t)} = f(x+(1+\varphi)\delta(x,t)),
	\end{equation} 
where $\varphi$ is the magnification factor.

The sample balancing process of the ME dataset is shown in Table \ref{table:equalization}. It is evident that the original ME dataset suffers from significant class imbalance. For instance, in the CASME II dataset, the ``negative'' class contains 88 samples, while the ``surprise'' class has only 25 samples. Similarly, in the SAMM dataset, the ``negative'' class consists of 92 samples, while the ``disgust'' class contains just 15 samples. By applying different combinations of magnification factors, we achieved three key objectives: (i) significantly increased the sample size (from 442 samples to 4022 samples), (ii) effectively mitigated the class imbalance issue, and (iii) enhanced the visibility of facial muscle movements. Experimental results indicate that excessive magnification (especially beyond a factor of 15) leads to severe facial distortions and reduced image quality. Therefore, we set the optimal maximum magnification factor to $\varphi$ = 14 in this study. Subsequently, we trained the AFE on the augmented ME dataset to improve its ability to capture more advanced and abstract ME features.

\subsection{Model variants: MPFNet-P and MPFNet-C}
\label{sec: variants}

To provide a more comprehensive explanation of the operational mechanism of MPFNet, this section elaborates on two variants, MPFNet-P and MPFNet-C, to clearly introduce their network architecture and workflow.

\subsubsection{MPFNet-P}

\begin{figure*}[t]
    \centering
    \includegraphics[width=0.85\linewidth]{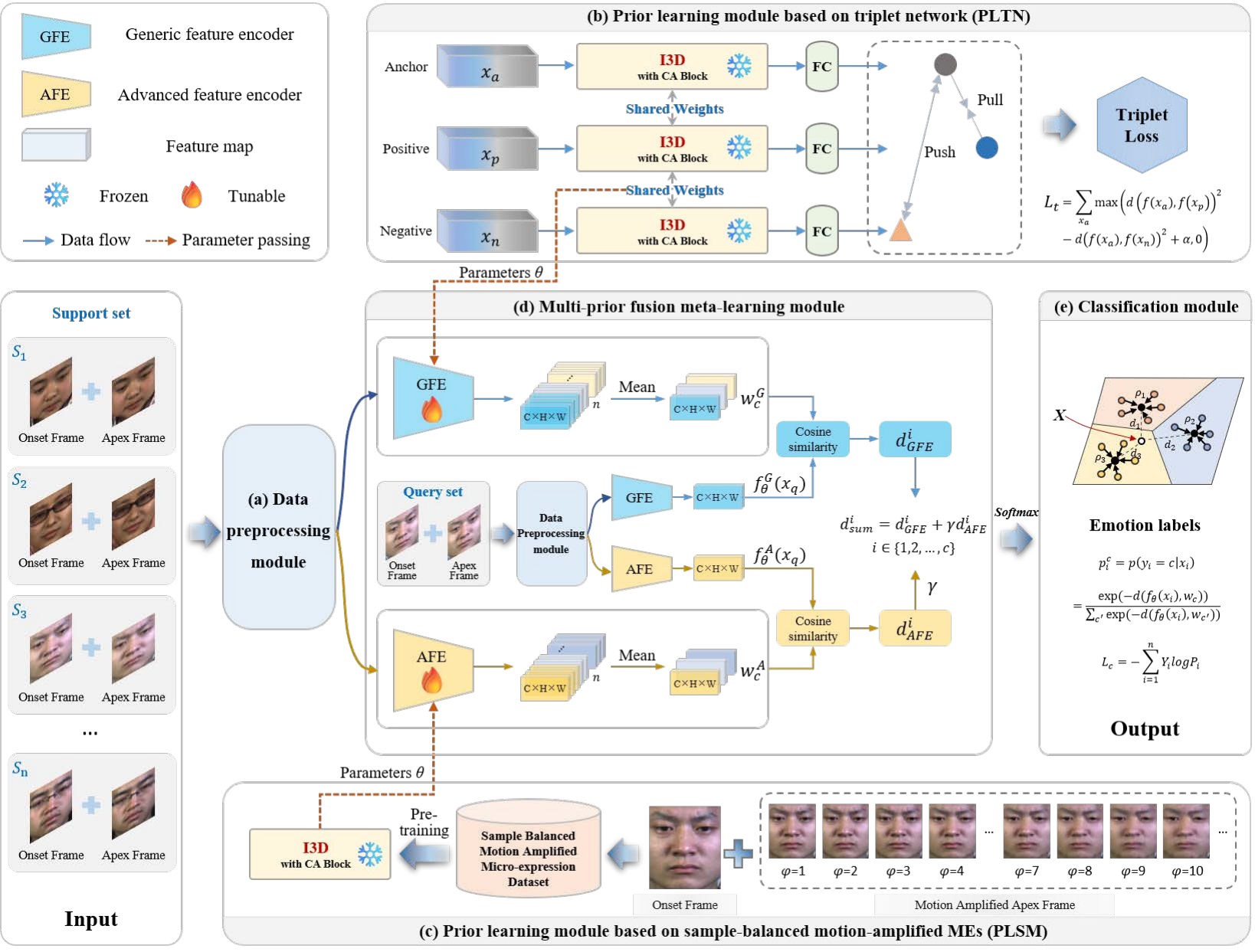}
    \caption{The structure of the MPFNet-P, which has a dual-stream architecture and includes five components. One data stream encodes the generic features of MEs and is represented by blue lines, while the other encodes advanced features and is represented by yellow lines. These dual streams are fused using a weighted-sum model fusion method for the final classification of MEs.}
    \label{figs:MPFNet-P}
\end{figure*}

After training the GFE and AFE, we integrate the encoders with complementary prior knowledge in parallel within a metric-based meta-learning framework, forming MPFNet-P. The architectural details of MPFNet-P are presented in Fig. \ref{figs:MPFNet-P}. This model adopts a parallel architecture design, where the GFE and AFE operate independently, extracting general and advanced features from the input data, respectively. This design enables the parallel processing of multi-perspective feature representations. The overall workflow of MPFNet-P is as follows:

(\romannumeral1) Data preprocessing module: The input to this module consists of image pairs formed by the onset and apex frames of ME samples. A frame interpolation method is applied to generate ME frame sequences of normalized length. Subsequently, optical flow and frame difference-based fusion features are computed to enhance the representation of spatiotemporal information. The position of this module within the workflow is illustrated in Fig. \ref{figs:MPFNet-P}(a), with further details provided in Section \ref{sec: Data preprocessing}.
(\romannumeral2) Prior learning based on a triplet network (PLTN): This module enables the model to learn generic feature extraction capabilities by constructing triplet ME samples and assessing their similarity. The acquired prior knowledge is stored in frozen parameters and subsequently used to initialize the GFE, as shown in Fig. \ref{figs:MPFNet-P}(b).
(\romannumeral3) Prior learning based on sample-balanced motion-amplified MEs (PLSM): This module involves training a CA-I3D model on a sample-balanced and augmented ME dataset to improve its ability to capture advanced ME features. Upon completion of training, the learned parameters are frozen and transferred to the AFE, as depicted in Fig. \ref{figs:MPFNet-P}(c). Additional details regarding these two prior learning processes can be found in Section \ref{sec: Pre-training}.

(\romannumeral4) Multi-prior fusion meta-learning module: Within the metric-based meta-learning framework, this module integrates the complementary prior knowledge of GFE and AFE to enhance the model’s MER capability, as illustrated in Fig. \ref{figs:MPFNet-P}(d). 
In this module, we first encode the deep feature vectors of the samples in both the support set $S$ and the query set $Q$. Then, we average the feature vectors of samples within the same class in the support set $S$ to obtain the mean feature vector, which serves as the class centroid in the embedding space. The specific calculation method is as follows:
\vspace{-0.2cm}

	\begin{equation}
		w_{c} = \frac{1}{|S_{c}|}\sum f_{\theta}(x_s),x_s\in S_{c},
        \label{wc}
	\end{equation}
where $f_{\theta }$ represents the feature encoder, $f_{\theta}(x_s)$ denotes the deep feature vector of a data sample $x_s$ in the support set $S$, $S_{c}$ is the sample cluster of the $c$-$th$ class, and $w_{c}$ is the centroid of the $c$-$th$ class.
Then, we employ a standard metric-based meta-learning process to compute the cosine similarity distance between each sample $f_{\theta }(x_q)$ in the query set $Q$ and the centroid of each class $w_c$ in the support set $S$ within the embedding space. The calculation is as follows:

    \begin{align}
        & d_{GFE}=similarity(f_{\theta }^{G}(x_q),w_c^G)=\frac{f_{\theta }^{G}(x_q) \cdot w_c^G}{\|f_{\theta }^{G}(x_q)\| \|w_c^G\|}, &\\
        & d_{AFE}=similarity(f_{\theta }^{A}(x_q),w_c^A)=\frac{f_{\theta }^{A}(x_q) \cdot w_c^A}{\|f_{\theta }^{A}(x_q)\| \|w_c^A\|}, &
    \end{align}
where $w_c^G$ and $w_c^A$ represent the centroids of each class in the support set $S$, computed from the deep features extracted using the GFE and the AFE, respectively. Similarly, $f_{\theta }^{G}(x_q)$ and $f_{\theta }^{A}(x_q)$ denote the deep feature vectors of the samples in the query set $Q$, encoded by the GFE and AFE, respectively. $d_{GFE}^{i}$ and $d_{AFE}^{i}$ denote the cosine similarity distances between features encoded by the GFE and AFE, respectively, and their corresponding centroids.

The two data streams are subsequently combined to enhance MER performance. Inspired by Gong \textit{et al.} \cite{gong2023meta}, the calculated distances of the dual-streams are added using a weighted-sum:

	\begin{equation}
		d_{sum}^{i} = d_{GFE}^{i}+\gamma d_{AFE}^{i},\ i\in \{ 1,2,...,c\},
	\end{equation}
where $d_{GFE}^{i}$ and $d_{AFE}^{i}$ denote the cosine similarity distances between the feature of the $i$-$th$ sample, encoded by GFE and AFE, respectively, and their corresponding centroids. The weighted distance $d_{sum}^{i}$ is then utilized to identify the nearest neighbour and predict the ME class label. Here, $\gamma$ represents the weighting coefficient, whose optimal value is determined through experimental validation as detailed in Section \ref{Results and analysis}.

(\romannumeral5) The classification module. The classification process involves computing the Euclidean distance between the feature representations of the query set and the centroid vectors of the support set. A nearest-neighbor approach is then employed to classify the samples in the query set, as illustrated in Fig. \ref{figs:MPFNet-P}(e). Further details are provided in Section \ref{sec: classification}.

\begin{figure*}[t]
    \centering
    \includegraphics[width=0.85\linewidth]{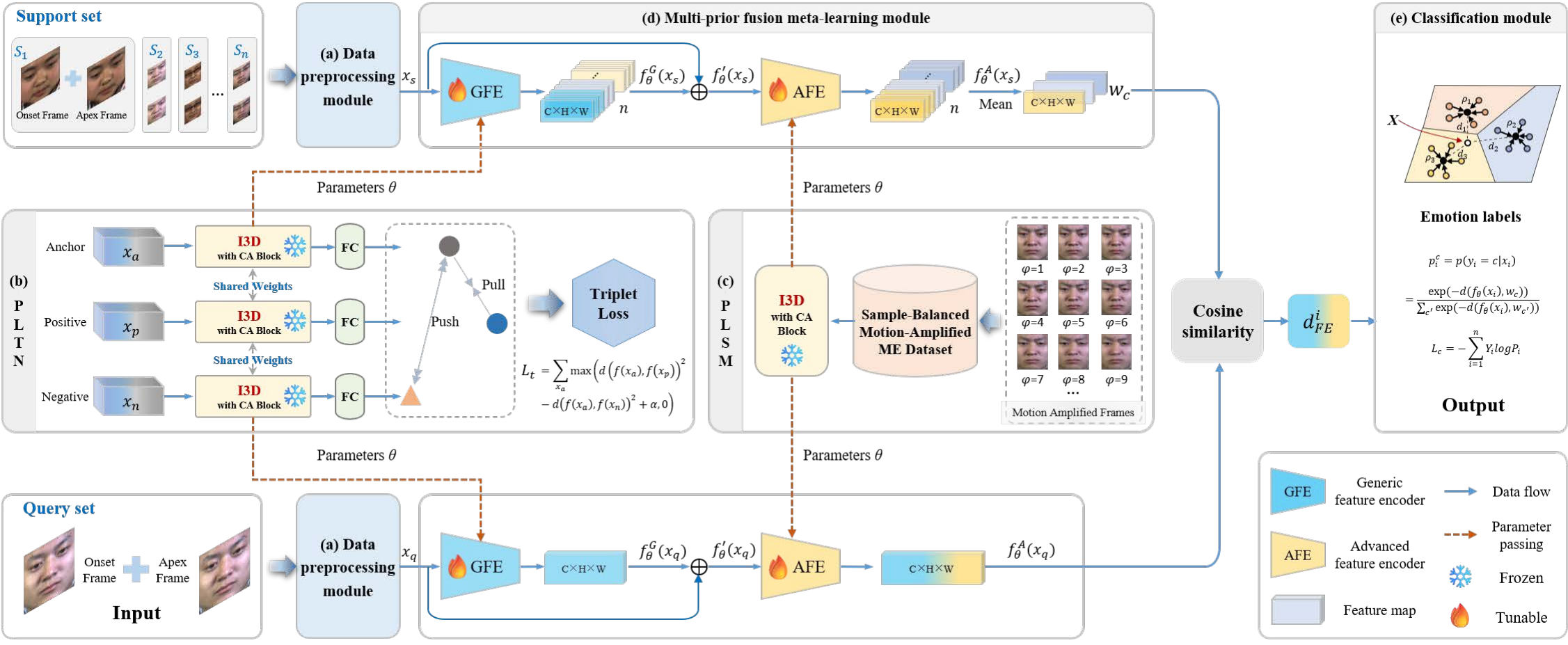}
    \caption{The architecture of MPFNet-C, which shares the same modules as MPFNet-P. The key distinction is that MPFNet-C employs a cascaded encoder architecture that progressively refines feature representations, enabling more effective capture of subtle and discriminative features.}
    \label{figs:MPFNet-C}
\end{figure*}

\subsubsection{MPFNet-C}

MPFNet-C retains the same five core modules as MPFNet-P. The primary distinction lies in its multi-prior fusion module, where the GFE and AFE are structured in a cascaded architecture. Furthermore, a residual structure is integrated to facilitate feature transmission and fusion, enhancing information flow across encoding layers, reducing potential feature loss, and improving representational capacity. The architectural details are illustrated in Fig. \ref{figs:MPFNet-C}.
In MPFNet-C, the support set samples undergo data preprocessing to obtain the initial features $x_s$. Subsequently, $x_s$ is encoded by the GFE, producing the general feature representation $f_{\theta}^{G}(x_{s})$ for MER. Next, we concatenate $f_{\theta}^{G}(x_{s})$ with the original feature $x_s$ along the channel dimension to form the intermediate feature representation $f_{\theta}^{\prime}(x_{s})$. This representation is then fed into the AFE for further encoding, yielding the advanced ME feature $f_{\theta}^{A}(x_{s})$. The process can be expressed as follows:

    \begin{align}
        & f_{\theta}^{G}(x_{s})=GFE(x_s), &\\
        & f_{\theta}^{\prime}(x_{s})=Concat(x_s, f_{\theta}^{G}(x_{s})), &\\
        & f_{\theta}^{A}(x_{s})=AFE(f_{\theta}^{\prime}(x_{s})).
    \end{align}

Next, for the feature vectors $f_{\theta}^{A}(x_{s})$ in the support set $S$, we compute the mean of features belonging to the same class to obtain the centroid $w_c$ for each class, as defined in Equation \ref{wc}. Similarly, the feature vectors $f_{\theta}^{G}(x_{q})$ of the samples in the query set $Q$ are computed in the same manner. The distance is then determined based on the cosine similarity between these vectors and the centroids of each class in the embedded space of the support set $S$, as follows:

    \begin{equation}
        d_{FE}=similarity(f_{\theta }^{A}(x_q),w_c)=\frac{f_{\theta }^{A}(x_q) \cdot w_c}{\|f_{\theta }^{A}(x_q)\| \|w_c\|}.
    \end{equation}



\subsection{Classification in meta-learning pipeline}
\label{sec: classification}

In this study, we consider MER as a few-shot classification problem and use a metric-based meta-learning framework to solve it. The few-shot MER problem is defined as follows, let $\tau $ denotes an $N$-way, $K$-shot few-shot learning task of MER from target domain, which consists of a labeled few-shot support set $S$ and unlabeled query set $Q$. $N$ denotes that the support set are from $N$ different classes, and $K$ is the number of labelled training samples in each class of a task. The query set data samples are also drawn from these $N$ categories, and the goal of an $N$-way, $K$-shot classification task is to classify unlabeled samples in the query set as one of the $N$ categories. We use a meta-learning pipeline based on a standard metrics to calculate the cosine similarity distance in embedding space between each data sample in query set $Q$ and the centroid of each class in support set $S$, and classify the samples in query set using a nearest neighbour method. The model is trained over many episodes to minimize the prediction error over the query set $Q$. In this paper, we set $K$ to 5 following the standard protocol of few-shot image classification problem.

In Section \ref{sec: variants}, we compute the distance between the deep feature vectors of the query set samples and the centroids of the clusters formed by the support set samples. The resulting cosine distances are then passed through a softmax function to calculate the probability that each sample $x_{i}$ belongs to class $c$. The calculation formula is as follows:

	\begin{equation}
		\mathcal{}{p_{i}^{c}} = p(y_{i}=c|x_{i}) =\frac{exp(-d(f_{\theta}(x_{i}),w_{c}))}{\sum exp(-d(f_{\theta}(x_{i}),w_{c^{'}})))},
	\end{equation}
where $d(f_{\theta}(x_{i}),w_{c}))$ represents the distance between the encoded feature of the query set, $f_{\theta}(x_{i})$, and the centroid vector $w_{c}$.
The classification loss function is defined as the cross-entropy loss between the predicted distribution and the ground truth distribution of the query set:

    \begin{equation}
        \mathcal{}{L_{c}} = -\sum_{i=1}^{n}Y_{i}\log{P_{i}},
    \end{equation}
where $P_{i}=\left [p_{1}^{i},p_{2}^{i},...,p_{c}^{i}  \right]$ is the predicted distribution, $C$ is the total number of ME categories, and $Y_{i}=\left [y_{1}^{i},y_{2}^{i},...,y_{c}^{i} \right]$ is the ground truth distribution of the $i$-$th$ data sample.


\vspace{-0.3cm}
\section{Experiments}
\label{experiments}

In this Section, we first provide a detailed description of the publicly available ME datasets used in this study. We then explain the classification tasks and evaluation metrics, followed by a discussion of the implementation details for model training and optimization.

\vspace{-0.2cm}
\subsection{ME datasets}
We employ three public ME datasets for experimental evaluation: SMIC \cite{pfister2011recognising}, CASME II \cite{yan2014casme} and SAMM \cite{davison2016samm}, along with their composite dataset, MEGC2019-CD \cite{see2019megc}. Below, we provide detailed characteristics of each dataset.

\textbf{SMIC} : There are 164 ME clips from 16 different subjects at 100 fps in SMIC, with 3 ethnicities. The resolution of samples is 640$\times$480 pixels. There are three ME types in SMIC, including negative, positive and surprise.

\textbf{CASME II}: The CASME II dataset contains 256 MEs samples from 26 subjects at 200 fps. There solution of the samples are 640$\times$480 pixels. The samples in CASME II are categorized into five ME classes, including happiness, surprise, disgust, repression and others.

\textbf{SAMM}: The SAMM dataset contains 159 ME instances from 32 participants at 200 fps and the resolution of the samples are 2,040$\times$1,088 pixels. The samples in SAMM demonstrates seven ME classes including happiness, surprise, disgust, repression, angry, fear and contempt.

\textbf{MEGC2019-CD}: The MEGC2019-CD dataset was introduced by the Micro-Expression Grand Challenge 2019 (MEGC2019) \cite{see2019megc}. It integrates three ME datasets—SMIC-HS, CASME II, and SAMM—and categorizes emotions into three groups: Negative (comprising ``Repression,'' ``Anger,'' ``Contempt,'' ``Disgust,'' ``Fear,'' and ``Sadness''), Positive (``Happiness''), and Surprise (``Surprise'').


\vspace{-0.1cm}
\subsection{Tasks and metrics}

Referring to previous research and the MEGC 2019, we conducted comprehensive experiments on the SMIC, CASME II, and SAMM datasets, including Single Database Evaluation (SDE) and Composite Database Evaluation (CDE).

\subsubsection{The SDE task}

The SDE task involves conducting experiments on each of the three datasets using their original emotion labels. Specifically, the SMIC dataset contains three emotion categories, while the CASME II and SAMM datasets share a common set of five emotion categories.

\subsubsection{The CDE task}

In the MEGC2019, the original emotion labels of the three datasets were consolidated into three broad categories: negative, positive, and surprise. Following this scheme, we first conducted three-class classification experiments separately on each dataset. Subsequently, to achieve a more comprehensive evaluation, we merged the three datasets into a composite dataset, MEGC2019-CD, and performed further experimental analyses.

\subsubsection{Evaluation metrics}

To independently evaluate each participant's samples, we employ the Leave-One-Subject-Out (LOSO) cross-validation method to assess MPFNet's performance. During each iteration, the model was optimized solely on the training set (i.e., all data except that of the current test subject), and hyperparameter tuning was conducted only on a subset of the training data (i.e., the validation set). Importantly, the test subject’s data was never involved in the hyperparameter tuning process. As a result, the final test outcomes remain unaffected by the tuning process, ensuring the validity and reliability of our evaluation. The training process of both the GFE and AFE was also conducted using LOSO cross-validation.
For the SDE task, we use accuracy (Acc) and F1-score (F1) for evaluation, with the F1-score providing a more objective and persuasive measure due to its robustness to class imbalance, especially in the CASME II and SAMM datasets. For the CDE task, we follow MEGC 2019 and use the unweighted F1-score (UF1) and unweighted average recall (UAR) to assess model performance. In fact, UF1 is commonly referred to as the macro-averaged F1-score, while UAR represents the ``balanced accuracy". The calculation methods for these metrics are detailed as follows:

    \begin{equation}
		\mathcal{}{Acc_{c}} = \frac{TP_{c}}{N_{c}},
    \end{equation}
 
    \begin{equation}
		\mathcal{}{F1_{c}} = \frac{TP_{c}}{2TP_{c}+FP_{c}+FN_{c}},
    \end{equation}

    \begin{equation}
		\mathcal{}{UF1} = \frac{1}{C} \sum_{c=1}^{C} F1_{c},
    \end{equation}
 
    \begin{equation}
		\mathcal{}{UAR} = \frac{1}{C} \sum_{c=1}^{C} Acc_{c},
    \end{equation}
where $TP_{c}$, $FP_{c}$, and $FN_{c}$ are the numbers of true positives, false positives, and false negatives for the $c$-$th$ class, respectively. $N_{c}$ is the number of samples in the $c$-$th$ class.

\subsection{Implementation details}

In the prior learning stage based on triplet network, we employed an SGD optimizer with a momentum of 0.9 and a learning rate of 0.01. Training was conducted for 60 epochs with a batch size of 128 and a weight decay of 5$\times$$10^{-4}$, resulting in a trained GFE. For the amplified-MEs based prior learning stage, we implemented a motion amplification algorithm to enhance the intensity of subtle movements in ME videos. This phase involves training for 80 epochs with an initial learning rate of 0.001, which was reduced by a factor of ten every 10 epochs. The SGD optimizer, with a momentum of 0.9 and a weight decay of 5$\times$$10^{-4}$, was maintained, resulting in a trained AFE. Within the meta-learning framework, we adopted an episode-based training strategy to enhance the model's generalization ability under few-shot conditions. Specifically, we designed two few-shot learning configurations for MER: 3-way 5-shot and 5-way 5-shot, which correspond to three-class and five-class classification scenarios, respectively. 
The query set samples are then randomly selected from the remaining samples of each category, ensuring no overlap with the support set samples. During training, we utilize the SGD optimizer with a fixed learning rate of 0.05 and a momentum parameter of 0.9. Each training batch contains 4 episodes, with the loss for each task calculated and averaged for gradient updates. All experiments were implemented using PyTorch and executed on an NVIDIA RTX 4090 GPU.

    \begin{table*}[b]
        \centering
        \caption{Comparison of MER Performance on the SDE task across different algorithms. The best results are highlighted in bold and the second best results are marked by underline. ``--'' denotes the results are not reported} 
        \begin{tabular}{p{3.6cm} p{1.6cm} p{1.6cm} p{1.6cm} p{1.6cm} p{1.6cm} p{1.6cm}}
            \hline
            \toprule[0.5pt]
            \multirow{2}{*}{Method} & \multicolumn{2}{c}{SMIC (3-class)} & \multicolumn{2}{c}{CASME II (5-class)} & \multicolumn{2}{c}{SAMM (5-class)} \\
            \cmidrule(r){2-3} \cmidrule(r){4-5} \cmidrule(r){6-7}
            & Acc & F1 & Acc & F1 & Acc & F1 \\ \hline
            LBP-TOP (2007) \cite{zhao2007dynamic} & 0.536 & 0.538 & 0.464 & 0.424 & -- & -- \\
            DiSTLBP-RIP (2017) \cite{huang2017discriminative} & 0.634 & -- & 0.647 & -- & -- & -- \\
            Bi-WOOF (2018) \cite{liong2018less} & 0.593 & 0.620 & 0.589 & 0.610 & 0.598 & 0.591 \\
            Micro-attention (2020) \cite{wang2020micro} & 0.494 & 0.496 & 0.659 & 0.539 & 0.485 & 0.402 \\
            GEME (2021) \cite{nie2021geme} & 0.646 & 0.616 & 0.752 & 0.735 & 0.558 & 0.454 \\
            MERSiamC3D (2021) \cite{zhao2021two} & -- & -- & 0.818 & \underline{0.830} & 0.687 & 0.640 \\
            FeatRef (2022) \cite{zhou2022feature} & 0.579 & -- & 0.628 & -- & 0.601 & -- \\
            Res-CapsNet (2023) \cite{shu2023res} & 0.756 & 0.749 & 0.763 & 0.736 & 0.683 & 0.543 \\
            SSRLTS-ViT (2024) \cite{zhang2024facial} & -- & -- & 0.746 & 0.736 & \underline{0.716} & \underline{0.716} \\ \hline
            MPFNet-P (Ours) & \underline{0.787} & \underline{0.787} & \underline{0.820} & 0.819 & 0.704 & 0.695 \\
            MPFNet-C (Ours) & \textbf{0.811} & \textbf{0.811} & \textbf{0.831} & \textbf{0.833} & \textbf{0.719} & \textbf{0.718} \\ \hline
            \toprule[0.5pt]
        \end{tabular}
        \label{table:single_dataset}
    \end{table*}

    \begin{table*}[b]
        \centering
        \caption{Comparison of MER Performance on the CDE task across different algorithms (3-class). The best results are highlighted in bold and the second best results are marked by underline}
        \begin{tabular}{p{3.5cm} p{1.1cm} p{1.1cm} p{1.1cm} p{1.1cm} p{1.1cm} p{1.1cm} p{1.1cm} p{1.1cm}}
            \hline
            \toprule[0.5pt]
            \multirow{2}{*}{Method} & \multicolumn{2}{c}{SMIC} & \multicolumn{2}{c}{CASME II} & \multicolumn{2}{c}{SAMM} & \multicolumn{2}{c}{MEGC2019-CD} \\
            \cmidrule(r){2-3} \cmidrule(r){4-5} \cmidrule(r){6-7} \cmidrule(r){8-9}
            & UF1 & UAR & UF1 & UAR & UF1 & UAR & UF1 & UAR \\ \hline
            LBP-TOP (2017) \cite{zhao2007dynamic} & 0.200 & 0.528 & 0.703 & 0.743 & 0.395 & 0.410 & 0.588 & 0.579 \\
            Bi-WOOF (2018) \cite{liong2018less} & 0.573 & 0.583 & 0.781 & 0.803 & 0.521 & 0.513 & 0.629 & 0.623 \\
            CapsuleNet (2019) \cite{van2019capsulenet} & 0.582 & 0.587 & 0.707 & 0.702 & 0.621 & 0.598 & 0.652 & 0.651 \\
            STSTNet (2019) \cite{liong2019shallow} & 0.680 & 0.701 & 0.838 & 0.869 & 0.659 & 0.681 & 0.735 & 0.760 \\
            RCN-A (2020) \cite{xia2020revealing} & 0.633 & 0.644 & 0.851 & 0.812 & 0.760 & 0.672 & 0.743 & 0.719 \\
            MERSiamC3D (2021) \cite{zhao2021two} & 0.736 & 0.760 & 0.882 & 0.876 & 0.748 & 0.728 & 0.807 & 0.799 \\
            AU-GCN (2021) \cite{lei2021micro} & 0.719 & 0.721 & 0.879 & 0.871 & 0.775 & 0.789 & 0.791 & 0.793 \\
            FeatRef (2022) \cite{zhou2022feature} & 0.701 & 0.708 & 0.891 & 0.887 & 0.737 & 0.715 & 0.783 & 0.783 \\
            Res-CapsNet (2023) \cite{shu2023res} & 0.690 & 0.685 & 0.812 & 0.813 & 0.680 & 0.686 & 0.741 & 0.745 \\
            RNAS-MER (2023) \cite{verma2023rnas} & 0.744 & 0.762 & 0.898 & 0.907 & 0.788 & 0.823 & \underline{0.830} & \textbf{0.851} \\
            LAENet (2024) \cite{gan2024laenet} & 0.662 & 0.652 & \underline{0.910} & \underline{0.911} & 0.681 & 0.662 & 0.756 & 0.740 \\ 
            TFT (2024) \cite{wang2024two} & 0.741 & 0.718 & 0.907 & 0.909 & 0.709 & 0.656 & 0.814 & 0.801 \\\hline
            MPFNet-P (Ours) & \underline{0.781}  & \underline{0.783}  & 0.879 & 0.895 & \underline{0.790}  & \underline{0.835}  & 0.811 & 0.820 \\
            MPFNet-C (Ours) & \textbf{0.806}  & \textbf{0.809}  & \textbf{0.911} & \textbf{0.923} & \textbf{0.795}  & \textbf{0.839}  & \textbf{0.840} & \underline{0.846} \\ \hline
            \toprule[0.5pt]
        \end{tabular}
        \label{table:combined_datasets}
    \end{table*}

\section{Results and analysis}
\label{Results and analysis}

In this Section, we first evaluate the performance of MPFNet on the SDE and CDE tasks. Next, we conduct ablation experiments to analyze the contributions of different prior knowledge and visual features. We then examine the impact of hyperparameter settings on model performance. Finally, we validate the effectiveness of the multi-prior learning strategy through visual analysis.

\subsection{Results of the SDE task}

For the SDE task, we conduct a comparative analysis of our MPFNet against several established methods for MER. The comparison encompasses both traditional hand-crafted feature-based approaches, including LBP-TOP \cite{zhao2007dynamic}, DiSTLBP-RIP \cite{huang2017discriminative}, and Bi-WOOF \cite{liong2018less}, as well as deep learning methods such as Micro-attention \cite{wang2020micro}, GEME \cite{nie2021geme}, MERSiamC3D \cite{zhao2021two}, FeatRef \cite{zhou2022feature}, RES-CapsNet \cite{shu2023res}, and SSRLTS-ViT \cite{zhang2024facial}. The experimental results, presented in Table \ref{table:single_dataset}, indicate the best-performing methods in bold and the second-best with underlining. It can be observed that MPFNet-C achieves the best performance, attaining the highest accuracy and F1 score.

\textbf{Comparison with prior learning-based methods.} Experimental results demonstrate that the multi-prior fusion strategy outperforms methods that rely on a single type of prior knowledge, such as GEME and MERSiamC3D. Compared to these two methods, both MPFNet-P and MPFNet-C demonstrate significant advantages in classification accuracy and F1 score across all datasets. For instance, on the SAMM dataset, MPFNet-C improves accuracy by 16.10\% compared to GEME and by 3.20\% compared to MERSiamC3D. Similarly, its F1-score surpasses that of GEME by 0.264 and MERSiamC3D by 0.078. We attribute this improvement to the fact that MPFNet integrates a more diverse and complementary set of complementary prior knowledge, while GEME relies solely on gender features as prior knowledge, and MERSiamC3D obtains prior knowledge by determining whether sample pairs are the same or different.

\textbf{Comparison with attention-based methods.} Our MPFNet also outperforms several attention-based methods, such as Micro-attention, FeatRef, and Res-CapsNet. For instance, compared to Res-CapsNet, which employs the ECA channel attention module \cite{wang2020eca}, our MPFNet-C achieves an accuracy improvement of 5.50\% on the SMIC dataset, 6.80\% on the CASME II dataset, and 3.60\% on the SAMM dataset. Additionally, MPFNet-C yields an increase of 0.062 in F1-score on SMIC, 0.097 on CASME II, and 0.175 on SAMM. The observed improvement can be attributed to the proposed CA-I3D model, which effectively captures spatiotemporal features and channel-wise information simultaneously, thereby significantly enhancing MER performance.

    \begin{table*}[b]
        \centering
        \caption{Ablation study of the prior learning strategy across three datasets. The best results are highlighted in bold}
        \label{tab:ablation-prior}
        \begin{tabular}{p{3.2cm} p{0.62cm} p{0.62cm} p{0.62cm} p{0.62cm} p{0.62cm} p{0.62cm} p{0.62cm} p{0.62cm} p{0.62cm} p{0.62cm} p{0.62cm} p{0.62cm}}
        \hline
        \toprule[0.5pt]
        \multirow{2}{*}{\begin{tabular}[c]{@{}l@{}}Prior learning (PL)\\ strategy\end{tabular}} & \multicolumn{2}{c}{\begin{tabular}[c]{@{}c@{}}SMIC\\ (3-class)\end{tabular}} & \multicolumn{2}{c}{\begin{tabular}[c]{@{}c@{}}SMIC\\ (5-class)\end{tabular}} & \multicolumn{2}{c}{\begin{tabular}[c]{@{}c@{}}CASME II \\ (3-class)\end{tabular}} & \multicolumn{2}{c}{\begin{tabular}[c]{@{}c@{}}CASME II \\ (5-class)\end{tabular}} & \multicolumn{2}{c}{\begin{tabular}[c]{@{}c@{}}SAMM \\ (3-class)\end{tabular}} & \multicolumn{2}{c}{\begin{tabular}[c]{@{}c@{}}SAMM\\ (5-class)\end{tabular}} \\
        \cmidrule(r){2-3} \cmidrule(r){4-5} \cmidrule(r){6-7} \cmidrule(r){8-9} \cmidrule(r){10-11} \cmidrule(r){12-13}
         & \multicolumn{1}{c}{Acc} & \multicolumn{1}{c}{F1} & \multicolumn{1}{c}{Acc} & \multicolumn{1}{c}{F1} & \multicolumn{1}{c}{Acc} & \multicolumn{1}{c}{F1} & \multicolumn{1}{c}{Acc} & \multicolumn{1}{c}{F1} & \multicolumn{1}{c}{Acc} & \multicolumn{1}{c}{F1} & \multicolumn{1}{c}{Acc} & \multicolumn{1}{c}{F1} \\ \hline
        w/o PL & 0.579 & 0.578 & 0.423 & 0.421 & 0.621 & 0.625 & 0.584 & 0.577 & 0.548 & 0.568 & 0.414 & 0.401 \\
        PLTN & 0.646 & 0.645 & 0.493 & 0.504 & 0.745 & 0.750 & 0.686 & 0.687 & 0.706 & 0.720 & 0.638 & 0.641 \\
        PLSM & 0.701 & 0.701 & 0.563 & 0.571 & 0.834 & 0.838 & 0.739 & 0.737 & 0.759 & 0.771 & 0.665 & 0.664 \\
        MPFNet-P (keep all) & 0.787 & 0.787 & 0.649 & 0.631 & 0.897 & 0.898 & 0.820 & 0.819 & 0.850 & 0.856 & 0.704 & 0.695 \\
        MPFNet-C (keep all) & \textbf{0.811} & \textbf{0.811} & \textbf{0.663} & \textbf{0.652} & \textbf{0.924} & \textbf{0.925} & \textbf{0.835} & \textbf{0.833} & \textbf{0.857} & \textbf{0.863} & \textbf{0.719} & \textbf{0.721} \\ \hline
        \toprule[0.5pt]
        \end{tabular}%
    \end{table*}

\textbf{Comparison with keyframe-based methods.} We also compare our approach with several key-frame-based MER methods, including Micro-attention, Res-CapsNet, and SSRLTS-ViT. These methods primarily rely on optical flow information between the onset and apex frames of MEs for feature extraction. Experimental results demonstrate that our video sequence-based approach outperforms these key-frame-based methods. Specifically, MPFNet-P achieves a 5.50\% higher accuracy and a 0.062 improvement in F1-score compared to Res-CapsNet on the SMIC dataset. On the CASME II dataset, MPFNet-P improves accuracy by 6.80\% and F1-score by 0.097, while on the SAMM dataset, accuracy increases by 3.60\% and F1-score by 0.175. These performance gains can be attributed to the ability of video sequence-based methods to more comprehensively capture the continuous temporal dynamics and subtle facial motion variations of MEs, whereas key-frame-based methods may fail to retain such critical information. Additionally, these results further validate the effectiveness of the adopted frame interpolation algorithm, which reconstructs high-quality motion information of MEs, thereby enhancing overall recognition performance.

\subsection{Results of the CDE task}

This section further validates the effectiveness of MPFNet on the CDE task. We strictly follow MEGC 2019 and conduct a series of three-classification experiments on the SMIC, CASME II, SAMM datasets, and their composite dataset, MEGC2019-CD. We compare MPFNet with both traditional handcrafted methods, such as LBP-TOP \cite{zhao2007dynamic} and Bi-WOOF \cite{liong2018less}, and deep learning methods, such as CapsuleNet \cite{van2019capsulenet}, STSTNet \cite{liong2019shallow}, RCN-A \cite{xia2020revealing}, MERSiamC3D \cite{zhao2021two}, FeatRef \cite{zhou2022feature}, RES-CapsNet \cite{shu2023res}, RNAS-MER \cite{verma2023rnas}, LAENet \cite{gan2024laenet} and TFT \cite{wang2024two}. The experimental results are presented in Table \ref{table:combined_datasets}. It can be observed that our method achieves the highest UF1 and UAR scores on most datasets. Compared to state-of-the-art deep learning methods such as LAENet and TFT, MPFNet demonstrates the most significant performance improvements on the SMIC and SAMM datasets. Specifically, on the SMIC dataset, MPFNet-C outperforms LAENet by 0.144 in UF1 and 0.157 in UAR. Similarly, on the SAMM dataset, MPFNet-C achieves UF1 and UAR improvements of 0.144 and 0.177, respectively, over LAENet. However, on the MEGC2019-CD dataset, MPFNet-C demonstrates a slightly lower UAR compared to RNAS-MER. This performance discrepancy may be attributed to the fact that RNAS-MER is specifically optimized for the three-class classification task of the MEGC2019-CD dataset, whereas our model exhibits superior capability in learning fine-grained categories, with its advantages becoming more pronounced in five-class classification tasks. 

\subsection{Ablation study}
\label{Ablation Study}

To assess the effectiveness of the proposed multi-prior fusion strategy and visual features, including optical flow and frame difference, we conduct a series of ablation experiments on the SMIC, CASME II, and SAMM datasets.

\subsubsection{The effect of prior learning strategy}

    \begin{table*}[t]
    \centering
    \caption{Ablation study of visual features across three datasets. The best results are highlighted in bold}
    \label{table:ablation-visual}
    \begin{tabular}{p{3.2cm} p{0.62cm} p{0.62cm} p{0.62cm} p{0.62cm} p{0.62cm} p{0.62cm} p{0.62cm} p{0.62cm} p{0.62cm} p{0.62cm} p{0.62cm} p{0.62cm}}
    \hline
    \toprule[0.5pt]
    \multirow{2}{*}{\begin{tabular}[c]{@{}c@{}}Visual feature\end{tabular}} & \multicolumn{2}{c}{\begin{tabular}[c]{@{}c@{}}SMIC\\ (3-class)\end{tabular}} & \multicolumn{2}{c}{\begin{tabular}[c]{@{}c@{}}SMIC\\ (5-class)\end{tabular}} & \multicolumn{2}{c}{\begin{tabular}[c]{@{}c@{}}CASME II \\ (3-class)\end{tabular}} & \multicolumn{2}{c}{\begin{tabular}[c]{@{}c@{}}CASME II \\ (5-class)\end{tabular}} & \multicolumn{2}{c}{\begin{tabular}[c]{@{}c@{}}SAMM \\ (3-class)\end{tabular}} & \multicolumn{2}{c}{\begin{tabular}[c]{@{}c@{}}SAMM\\ (5-class)\end{tabular}} \\
    \cmidrule(r){2-3} \cmidrule(r){4-5} \cmidrule(r){6-7} \cmidrule(r){8-9} \cmidrule(r){10-11} \cmidrule(r){12-13}
     & \multicolumn{1}{c}{Acc} & \multicolumn{1}{c}{F1} & \multicolumn{1}{c}
     {Acc} & \multicolumn{1}{c}{F1} & \multicolumn{1}{c}{Acc} & \multicolumn{1}{c}{F1} & \multicolumn{1}{c}{Acc} & \multicolumn{1}{c}{F1} & \multicolumn{1}{c}{Acc} & \multicolumn{1}{c}{F1} & \multicolumn{1}{c}{Acc} & \multicolumn{1}{c}{F1} \\ \hline
    Optical flow & 0.731 & 0.732 & 0.593 & 0.591 & 0.834 & 0.838 & 0.769 & 0.767 & 0.788 & 0.780 & 0.665 & 0.662 \\
    Frame difference & 0.530 & 0.532 & 0.421 & 0.422 & 0.623 & 0.620 & 0.579 & 0.577 & 0.599 & 0.591 & 0.503 & 0.500 \\
    MPFNet-P (keep all) & \textbf{0.787} & \textbf{0.787} & \textbf{0.649} & \textbf{0.631} & \textbf{0.897} & \textbf{0.898} & \textbf{0.820} & \textbf{0.819} & \textbf{0.850} & \textbf{0.856} & \textbf{0.704} & \textbf{0.695} \\ \hline
    Optical flow & 0.752 & 0.750 & 0.614 & 0.615 & 0.865 & 0.866 & 0.782 & 0.783 & 0.799 & 0.800 & 0.678 & 0.680 \\
    Frame difference & 0.573 & 0.571 & 0.496 & 0.495 & 0.662 & 0.664 & 0.596 & 0.597 & 0.609 & 0.607 & 0.515 & 0.514 \\
    MPFNet-C (keep all) & \textbf{0.811} & \textbf{0.811} & \textbf{0.663} & \textbf{0.652} & \textbf{0.924} & \textbf{0.925} & \textbf{0.835} & \textbf{0.833} & \textbf{0.857} & \textbf{0.863} & \textbf{0.719} & \textbf{0.721} \\ \hline
    \toprule[0.5pt]
    \end{tabular}%
\end{table*}

We design multiple experimental conditions by progressively incorporating prior knowledge. These conditions include without prior learning (w/o PL), where the encoder is trained from scratch; prior learning based on triplet network (PLTN); prior learning based on sample-balanced motion-amplified MEs (PLSM); and the fusion of both prior learning methods within the MPFNet-P or MPFNet-C framework. The experimental results are presented in Table \ref{tab:ablation-prior}. It is evident that the model performance significantly improves with the gradual incorporation of prior knowledge, especially with the MPFNet model that adopts a multi-prior fusion strategy, which achieves the best performance. Furthermore, MPFNet-C outperforms MPFNet-P across all evaluation metrics, indicating that the cascade feature encoder structure is more effective for MER tasks than the parallel feature encoder structure.

\subsubsection{The effect of visual features}

Optical flow features serve as a crucial motion representation method, effectively capturing pixel-level motion information between video frames. Consequently, they have been widely applied in MER. Meanwhile, frame difference features, which quantify pixel intensity variations between consecutive frames, provide complementary visual cues. This study combines both features to form a comprehensive visual representation, as described in Section \ref{method}. Few studies have evaluated the relative contributions of these two visual features in MER tasks. To fill this gap, we conducted an ablation experiment to identify the dominant feature. We performed extensive experiments on both the MPFNet-P and MPFNet-C architectures across three feature configurations: (i) optical flow features only, (ii) frame difference features only, and (iii) integrated features combining both modalities. As demonstrated in Table \ref{table:ablation-visual}, our experimental results reveal two key findings: First, optical flow features consistently outperform frame difference features across both model architectures. Second, the feature fusion strategy yields significant performance improvements over single-feature approaches. Specifically, for the three-class classification task on the CASME II dataset, MPFNet-C achieves an accuracy of 0.865 with optical flow features alone, compared to 0.662 using only frame difference features. Notably, the integrated feature approach boosts accuracy to 0.924, demonstrating the complementary nature of these feature modalities. These results confirm the dominant role of optical flow features in MER tasks, while also demonstrating that incorporating frame difference features effectively improves the comprehensive representation capability of visual features. This study provides empirical evidence for understanding the contributions of different visual features in MER.

\subsection{Hyperparameter settings}

\begin{figure}[t]
    \centering
    \includegraphics[width=0.85\linewidth]{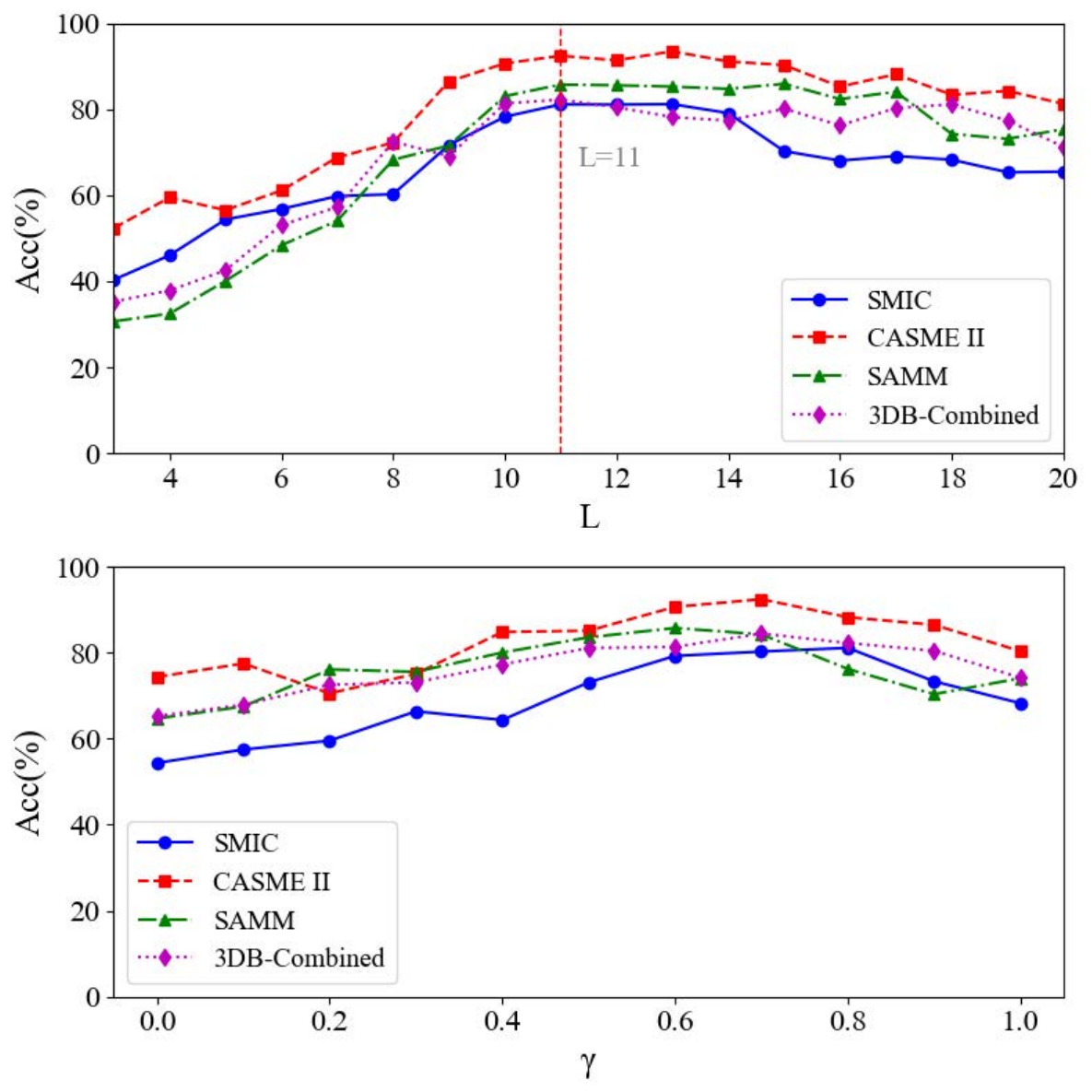}
    \caption{Impact of hyperparameters on MER: Frame sequence length $L$ (Top) and weighting factor $\gamma$ (Bottom).}
    \label{figs:hyperparameter}
\end{figure}

In this study, we introduce two critical hyperparameters: the length of the ME frame sequence ($L$) after interpolation and the distance-weighting factor ($\gamma $) in MPFNet-P. To identify their optimal values, we conducted a series of experiments, varying $L$ within the range $\{3,4,\dots,20\}$ and $\gamma$ within $\{0.0,0.1,\dots,1.0\}$. The results, presented in Fig. \ref{figs:hyperparameter}, demonstrate that the three-class classification accuracy peaks and stabilizes when $L$ ranges between 11 and 13 across all datasets. However, further increasing $L$ results in accuracy fluctuations or declines, likely due to information redundancy, noise accumulation, and heightened computational complexity, which collectively degrade classification performance. To achieve a balance between model performance and computational efficiency, we set $L=11$. As for $\gamma$, the optimal values for MPFNet-P were found to be 0.8, 0.7, 0.6, and 0.7 on the SMIC, CASME II, SAMM, and MEGC2019-CD datasets, respectively. These findings suggest that the AFE encoder plays a predominant role in feature representation learning within the embedding space.

\subsection{Visual analysis}
\label{visualization and analysis}

\begin{figure*}[t]
    \centering
    \includegraphics[width=0.70\linewidth]{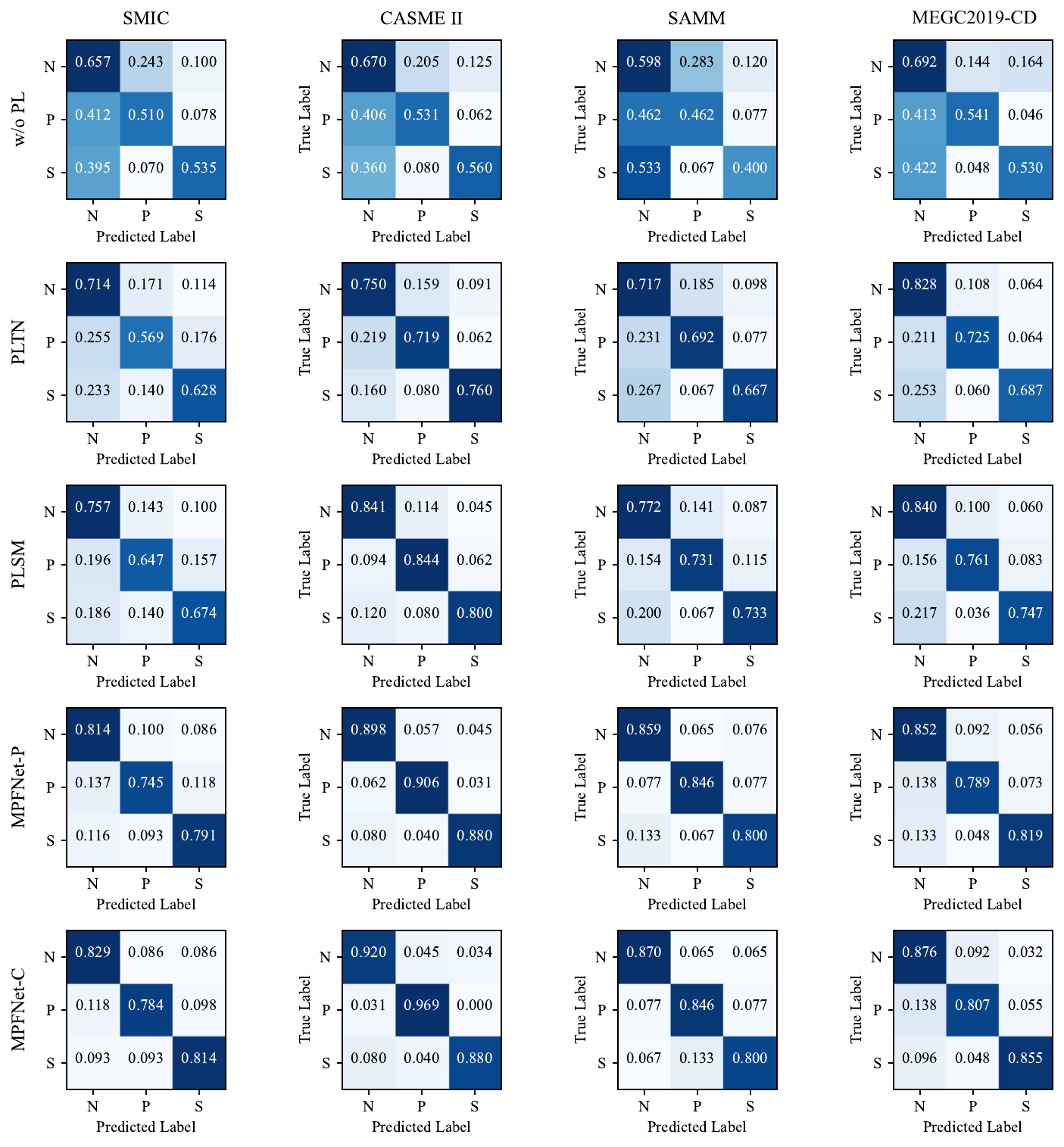}
    \caption{The confusion matrices for MER with different prior learning strategy on SMIC, CASME II, SAMM and the MEGC2019-CD datasets. The terms w/o PL, PLTN, PLSM, and MPFNet refer to four distinct prior learning strategies. N, P, and S stand for negative, positive, and surprise respectively.}
    \label{figs:confusion_matrices}
\end{figure*}

\textbf{Visualization of confusion matrices.} To gain further insight into the proposed method, we visualize the confusion matrices for different prior learning strategies across four datasets, as shown in Fig. \ref{figs:confusion_matrices}. The diagonal elements represent the proportion of correctly classified MEs in the test set, with darker colors indicating higher accuracy. It is evident that our model, when trained from scratch without prior knowledge, exhibits poor classification accuracy with significant variation across different categories. As prior knowledge is gradually introduced, we observe a substantial improvement in MPFNet's performance in recognizing positive, negative, and surprise expressions. Notably, the negative class contains the most samples across the three datasets, particularly in the CASME II and SAMM datasets. Many existing algorithms achieve high classification accuracy for this dominant category, often at the expense of reduced accuracy for the other two categories. The MPFNet proposed in this paper significantly improves the accuracy of the two secondary categories, achieving a more balanced accuracy distribution across all categories.
For instance, without the integration of prior knowledge, the standard deviations of accuracy for the three emotions in the SMIC, CASME II, SAMM, and MEGC2019-CD datasets are 0.079, 0.073, 0.101, and 0.090, respectively. After incorporating multiple sources of prior knowledge, these standard deviations are reduced to 0.022, 0.044, 0.035, and 0.035 on MPFNet-C. These results demonstrate that the multi-prior learning strategy designed in this study mitigates the impact of few-shot and imbalance issues on the accuracy of MER.

\begin{figure*}[t]
    \centering
    \includegraphics[width=0.74\linewidth]{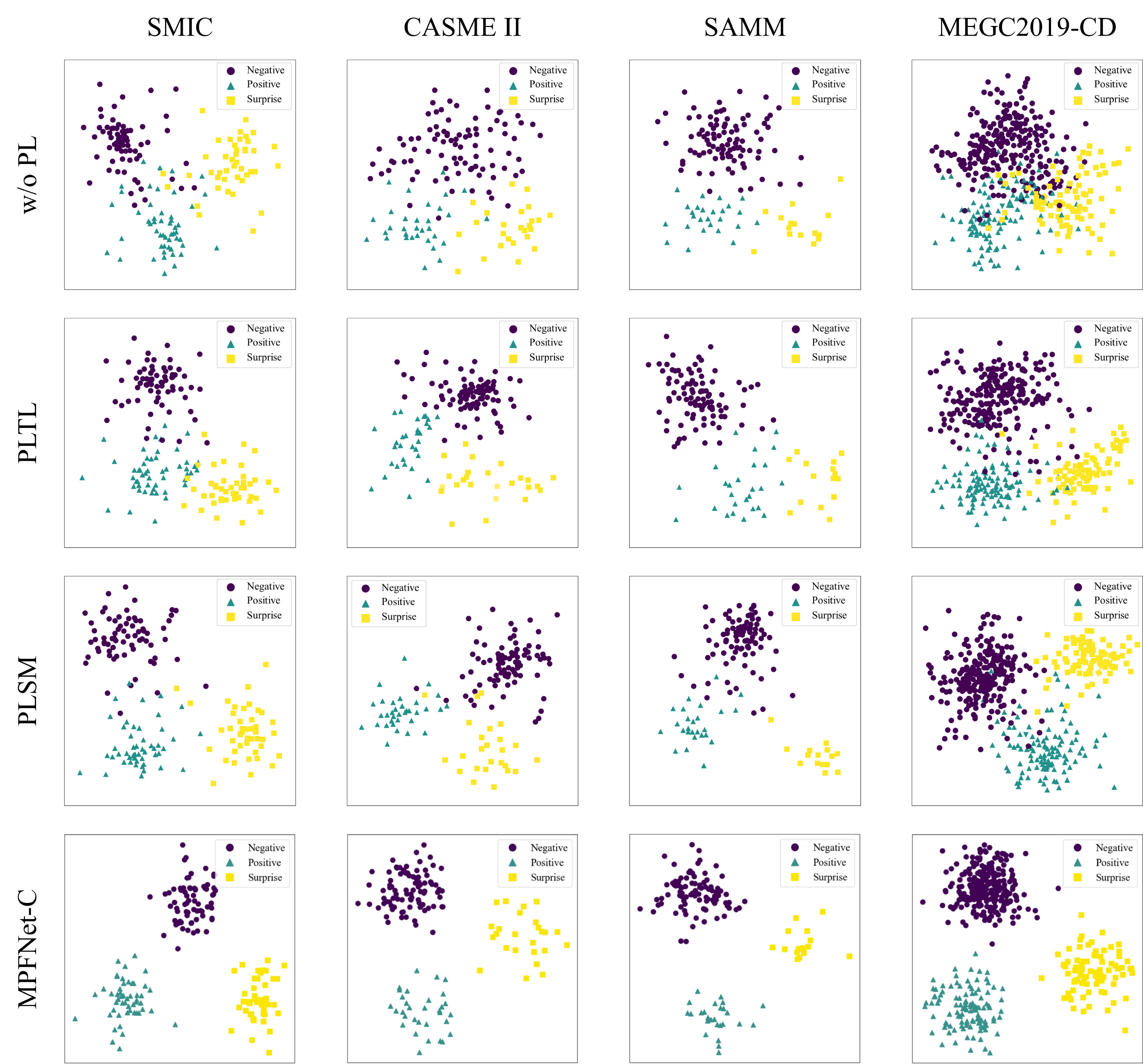}
    \caption{The t-SNE algorithm is utilized for visualizing deep features in the three-class classification task of negative, positive, and surprise expressions. The terms w/o PL, PLTN, and PLSM represent different prior learning strategies. As prior knowledge is progressively integrated, the boundaries between categories become increasingly distinct.}
    \label{figs:tSNE}
\end{figure*}

\textbf{Visualization of feature distribution.} We utilize the t-SNE method to project the feature distribution of the deep model into a two-dimensional space, visualizing it as a scatter plot. As shown in Fig. \ref{figs:tSNE}, the feature space extracted by the model without prior knowledge exhibits significant overlap, with samples from all three categories blending together and becoming indistinguishable. In contrast, when PLTN and PLSM are applied, the boundaries between categories become progressively wider and more distinct. After incorporating both types of prior knowledge, MPFNet-C learns more compact intra-class features, while the inter-class features for negative, positive, and surprise samples form tighter clusters with clearer boundaries, making them easier to separate. This demonstrates that our model extracts more discriminative features, resulting in tighter clusters that enhance the MER capability.

\begin{figure}[t]
    \centering
    \includegraphics[width=0.9\linewidth]{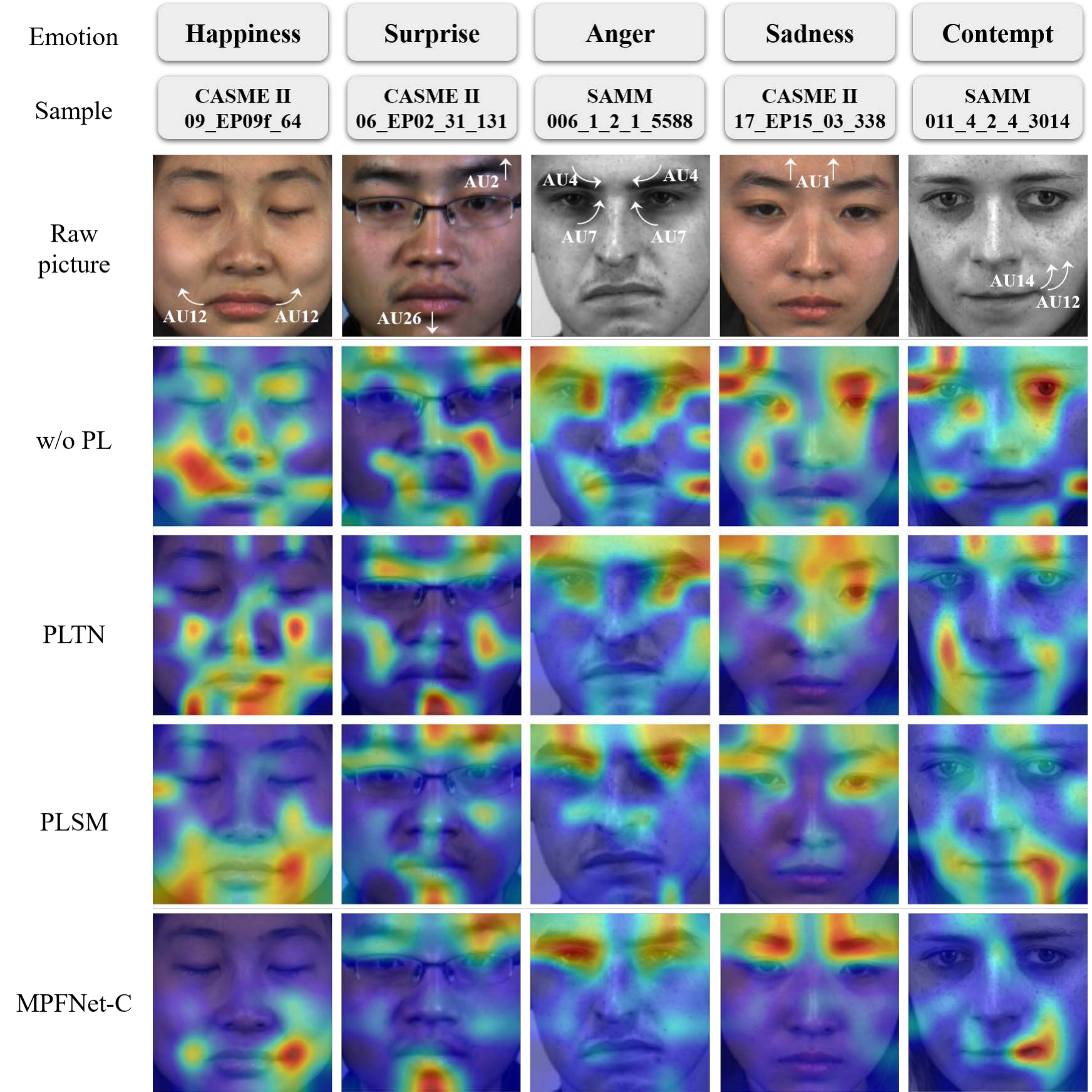}
    \caption{Visual explanations of ME via gradient-based localization. We select a sample from each of the five emotional categories, apply the Grad-CAM method after the last convolutional layer of the model, and then superimpose the generated heatmap on the original sample image.}
    \label{figs:grad_cam}
\end{figure}

\textbf{Visualization of feature heatmaps.} To gain a deeper understanding of the learned features, we visualize the activation heatmaps using Grad-CAM\footnote{\url{https://github.com/jacobgil/pytorch-grad-cam}}, as shown in Fig. \ref{figs:grad_cam}. This visualization illustrates the model's capability to identify distinct regional distributions and visual features in images. Grad-CAM generates localization maps highlighting regions activated during facial feature extraction. We select one sample from each of the five emotional categories, apply Grad-CAM after the final convolutional layer of the model, and superimpose the resulting heatmap onto the original sample image. Initially, the model without prior learning focuses on regions unrelated to MEs, negatively affecting its performance. After incorporating the multi-prior learning strategy, the highlighted regions gradually converge towards key facial areas—such as the eyebrows and corners of the mouth—that are critical for detecting subtle MEs. Specifically, for the ``happiness'' sample, the Grad-CAM heatmap highlights the zygomaticus major muscle, corresponding to AU12, with the action descriptor ``Lip corner puller.'' For the ``surprise'' sample, the highlighted regions include the frontalis (pars lateralis) and masseter muscles, corresponding to AU2 (``Outer brow raiser'') and AU26 (``Jaw drop''), respectively. For the ``anger'' sample, the heatmap highlights the corrugator supercilii and orbicularis oculi muscles, consistent with AU4 (``Brow lowerer'') and AU7 (``Lid tightener''), respectively. For the ``sadness'' sample, the highlighted region corresponds to the frontalis (pars medialis), associated with AU1 (``Inner brow raiser''). For the ``contempt'' sample, the zygomaticus major and zygomaticus minor muscles are highlighted, corresponding to AU12 ((``Lip corner puller'') and AU14 (``Dimpler''), respectively. These visualized heatmaps provide strong evidence of the model's effectiveness.
It should be noted that the MPFNet-P model employs a parallel fusion architecture design for feature encoders. This unique architectural approach renders both t-SNE and Grad-CAM techniques inapplicable to this model. Consequently, the visualization results pertaining to the MPFNet-P model are not included in Fig. \ref{figs:tSNE} and \ref{figs:grad_cam}.





\section{Conclusion}
\label{conclusion}

This paper proposes a multi-prior fusion network (MPFNet), offering an innovative approach to effectively utilize scarce ME data and address class imbalance issues. First, we design a prior learning strategy based on a triplet network to train the model for encoding general ME features. To overcome the limitations of ME samples in transfer learning, we construct a sample-balanced and motion-amplified ME dataset to further train the model and extract more advanced ME features. Both feature encoders adopt the CA-I3D model as the backbone, enabling the efficient learning of crucial spatiotemporal and channel features. Furthermore, we designed two model variants, MPFNet-P and MPFNet-C, to evaluate the impact of different prior knowledge integration strategies on MER. Experimental results demonstrate that the proposed method not only improves the overall classification accuracy of ME recognition but also ensures balanced performance across different categories. Future research will focus on multimodal ME datasets, such as CAS(ME)$^3$ \cite{li2022cas}, to further explore multimodal ME features and develop more efficient fusion strategies. The ultimate goal is to achieve a more robust and generalized MER framework.

\section{Ethical impact statement}
\label{Ethical}

Privacy and data protection are paramount in ME research. ME data may contain sensitive biometric information, and deep learning models could potentially identify specific patterns from such data. Therefore, it is crucial to safeguard both the original data and the learned patterns. This requires secure model storage and the implementation of robust privacy-preserving techniques to prevent sensitive information leakage. The public ME dataset used in this study was collected with informed consent from participants, covering aspects such as data collection, processing, and sharing. Additionally, the optical flow and frame-difference extraction methods applied in this study effectively eliminate sensitive information, such as appearance and gender, while preserving the facial motion characteristics essential for ME analysis. This approach ensures the ethical development and deployment of MER systems. 


\section*{Acknowledgments}
This work was supported in part by the grants from the National Natural Science Foundation of China under Grant (No.62332019, No.62076250, and No.62406338), the National Key Research and Development Program of China (No.2023YFF1203900 and No.2023YFF1203903).




\bibliographystyle{IEEEtran}
\bibliography{ref_supp}

\end{document}